\documentclass{article}

\usepackage[main, final]{neurips_2026}

\usepackage[utf8]{inputenc} 
\usepackage[T1]{fontenc}    
\usepackage{hyperref}       
\usepackage{url}            
\usepackage{booktabs}       
\usepackage{amsfonts}       
\usepackage{nicefrac}       
\usepackage{microtype}      
\usepackage{xcolor}         

\usepackage{amsmath}
\usepackage{amssymb}
\usepackage{mathtools}
\usepackage{amsthm}

\usepackage{placeins}

\usepackage{algorithm}
\usepackage{algpseudocode}
\usepackage{booktabs}
\usepackage{enumitem}
\usepackage{subcaption}
\usepackage{multirow}
\usepackage{amsmath}
\usepackage[most]{tcolorbox}
\usepackage{xcolor}
\usepackage{caption}         
\usepackage{wrapfig}

\definecolor{algBg}{HTML}{F9FAFB}        
\definecolor{algBorder}{HTML}{D8DEE4}    
\definecolor{algAccent}{HTML}{2563EB}    
\definecolor{algAccentLite}{HTML}{EAF2FF}
\definecolor{algTitle}{HTML}{0F172A}     
\definecolor{algComment}{HTML}{5B6776}   
\definecolor{algLine}{HTML}{A8B3C2}      

\tcbset{enhanced}
\newtcolorbox{AlgWrap}{
  colback=algBg,
  colframe=algBorder,
  boxrule=0.35mm,
  arc=2mm,
  left=3.0mm, right=3.0mm, top=2.0mm, bottom=2.0mm
}

\algrenewcommand\alglinenumber[1]{\scriptsize\color{algLine}#1}
\algrenewcommand\algorithmiccomment[1]{\hfill{\scriptsize\color{algComment}\(\triangleright\)~#1}}
\algrenewcommand\algorithmicrequire{\textbf{\color{algAccent}Require:}}
\algrenewcommand\algorithmicensure{\textbf{\color{algAccent}Return:}}
\algrenewcommand\algorithmiccomment[1]{\hfill{\(\triangleright\)~#1}}

\newtcbox{\AlgSectionBox}[1][]{%
  on line,
  colback=algAccentLite,
  colframe=algAccent,
  boxrule=0.3mm,
  arc=1mm,
  left=1mm, right=1mm, top=0.2mm, bottom=0.2mm,
  nobeforeafter,#1}

\usepackage{tikz}
\usetikzlibrary{arrows.meta,positioning,fit,calc,backgrounds}
\newcommand{\AlgSection}[1]
{\AlgSectionBox{\textbf{\color{algAccent}#1}}}
\makeatletter
\renewcommand{\ALG@step}{\refstepcounter{ALG@line}}
\makeatother

\usepackage[capitalize,noabbrev]{cleveref}

\theoremstyle{plain}

\theoremstyle{definition}

\theoremstyle{remark}

\usepackage[textsize=tiny]{todonotes}

\title{PDE-SSM: A Spectral State Space Approach to Spatial Mixing in Diffusion Transformers}

\author{%
  Eshed Gal \\
  University of British Columbia\\
  Vancouver, BC, Canada \\
  \texttt{eshedg@cs.ubc.ca} \\
  \And
  Moshe Eliasof \\
  Ben-Gurion University of the Negev \\
    Beersheba, Israel \\
  \texttt{eliasof@bgu.ac.il} \\
  \And
  Eldad Haber \\
University of British Columbia\\
  Vancouver, BC, Canada \\
  \texttt{ehaber@eoas.ubc.ca} \\
}

\begin{document}

\maketitle

\begin{abstract}
The success of vision transformers—especially for generative modeling—is limited by the quadratic cost and weak spatial inductive bias of self-attention. We propose PDE-SSM, a spatial state-space block that replaces attention with a learnable convection–diffusion–reaction partial differential equation. This operator encodes a strong spatial prior by modeling information flow via physically grounded dynamics rather than all-to-all token interactions. Solving the PDE in the Fourier domain yields global coupling with near-linear complexity,
delivering a principled and scalable alternative to attention. We integrate PDE-SSM into a flow-matching generative model to obtain the PDE-based Diffusion Transformer PDE-SSM-DiT. Empirically, PDE-SSM-DiT matches or exceeds the performance of 
matched Diffusion Transformers while substantially reducing compute. Our results show that, analogous to 1D settings where SSMs supplant attention, multi-dimensional PDE operators provide an efficient, inductive-bias-rich foundation for next-generation vision models.
\end{abstract}

\section{Introduction}

The Transformer architecture \citep{vaswani2017attention}, with its self-attention mechanism, has become the de facto standard for state-of-the-art machine learning techniques, extending its dominance from Natural Language Processing \citep{ touvron2023llama} to Computer Vision \citep{dosovitskiy2020image, peebles2023scalable}. In generative modeling for images, the U-Net \citep{ronneberger2015u} backbone with attention mechanism has become the standard for diffusion models over the years \citep{ho2020denoising,song2020denoising,nichol2021improved,dhariwal2021diffusion,rombach2022high,lipman2023flow}. 
However, after a long streak of success, the Diffusion Transformer (DiT) \citep{peebles2023scalable} marked a significant milestone by replacing the U-Net backbone with a Transformer-based architecture, demonstrating that pure Transformer models can achieve or surpass the performance of convolutional U-Net backbones in large-scale diffusion models.
Like other transformers, DiTs utilize deep networks.
However, it is worth noting that DiTs are also constructed using attention mechanisms, which makes them fundamentally scalable.
Be it modern U-Nets or DiTs, these are attention-based models which essentially operate by treating an image as a collection of patches and apply full weight within the patch and self-attention to model the relationship between patches \citep{dosovitskiy2020image, liu2021swin, peebles2023scalable}.

Despite its empirical success, self-attention has two well-known drawbacks in the spatial domain \citep{tay2022efficienttransformerssurvey}, which are the focus of this paper. The limitations of self-attention in this case are as follows: \emph{first}, its computational and memory requirements become increasingly challenging in 2D inputs, which have quadratically as many tokens as standard 1D inputs. Although there are a few recent ideas to attempt and reduce this cost (e.g. linear attention) \citep{katharopoulos2020transformers, shen2018efficient}, their performance generally falls short compared to standard attention-based methods. 
\emph{Second}, self-attention is fundamentally permutation-invariant, meaning it does not inherently encode spatial locality or grid structure. Positional encodings are commonly introduced to remedy this limitation, and indeed, they allow Transformers to capture ordering and relative spatial information. Nevertheless, this mechanism is not as direct or specialized as the inductive bias present in convolutional neural networks (CNNs), which are explicitly designed around locality. Consequently, Transformer-based models often rely more heavily on data to learn robust spatial relationships.

Concurrently, a distinct lineage of sequence models has emerged to address these limitations in the 1D domain. State Space Models (SSMs), particularly structured SSMs (S4) \citep{gu2021efficiently} and their modern successors like Mamba \citep{gu2023mamba}, have achieved remarkable success on long-sequence tasks while offering compelling efficiency. These models are grounded in the theory of continuous-time systems, governed by a linear Ordinary Differential Equation (ODE):
\begin{equation}
    \label{eq:ssm}
    \frac{dh(t)}{dt} = Ah(t) + Bu(t).
\end{equation}
By imposing structure on the state matrix $A$, these models can be formulated as a recurrent or convolutional system, enabling them to capture extremely long-range dependencies in linear or near-linear time \citep{gu2021efficiently}.  Their success raises a critical and compelling question: \emph{how can the principles of SSMs be generalized from 1D time to N-D  space to create a more efficient and spatially-aware foundation for vision models?}

Recent work has extended SSMs to graphs \citep{ceni2025message,lahoti2025chimera},  replacing $A$ with the adjacency matrix. 
In the context of SSMs in 2D, initial forays into this area, such as Vision Mamba (ViM) \citep{zhu2024vision}, have adapted 1D SSMs to images by "flattening" the 2D image domain into a 1D sequence, followed by applying the SSM model along forward and backward paths.  
While effective, this approach imposes an artificial causality on the spatial domain, effectively processing pixels/nodes in a scan order rather than respecting the true multi-dimensional structure of images.
In this sense, it represents a heuristic adaptation of 1D principles, rather than a full generalization to the spatial setting. We provide an additional discussion of concurrent and complementary approaches in Appendix~\ref{sec:additional_related_work}.

\paragraph{Our approach.} To bridge this gap, in this work, we propose a more principled approach. Concretely, we argue that the natural continuous-space generalization of an ODE that is the core of SSMs as shown in \Cref{eq:ssm}, is a Partial Differential Equation (PDE). 
This observation has been studied for the control of PDEs in the context of chemical engineering \citep{morris2010control, foias1996robust}.

We extend this idea and introduce the \emph{PDE State-Space Model (PDE-SSM)}, an architectural block that models the evolution of a hidden state $h(t, \mathbf{x})$ in time $t$, where $\mathbf{x} \in \mathbb{R}^d$ --- according to a learnable, general-form diffusion–convection–reaction equation, as follows:
\begin{equation}
    \label{eq:ssm_nd}
    \frac{\partial h}{\partial t} = \underbrace{\nabla \cdot (K \nabla h)}_{\text{Diffusion}} + \underbrace{\mathbf{b} \cdot \nabla h}_{\text{Convection}} + \underbrace{rh}_{\text{Reaction}},
\end{equation}
where the diffusion tensor $K$, convection vector field $\mathbf{b}$, and reaction term $r$ are learnable parameters. This formulation provides a flexible and physically-grounded inductive bias; information propagates across the spatial domain via structured and well-studied mechanisms: (i) diffusion smooths and aggregates local information; (ii) convection directs the flow of information; and (iii) reaction models local feature transformations. 
Importantly, by implementing the solution to the PDE in \Cref{eq:ssm_nd} in the Fourier domain, the underlying operator reduces the spatial mixing complexity to $\mathcal{O}(N \log N)$. While channel coupling introduces an $\mathcal{O}(N C_{\rm hid}^2)$ cost, the overall near-linear scaling with respect to sequence length makes it highly scalable relative to the quadratic $\mathcal{O}(N^2)$ spatial cost of self-attention, while still capturing global interactions more directly than local convolutions. We integrate the PDE-SSM block into a standard transformer architecture, replacing the self-attention layers to create the PDE-based Diffusion Transformer, to obtain PDE-SSM\textsubscript{DiT}.

\paragraph{Our contributions are as follows:}
\begin{enumerate}[leftmargin=2em]
    \item We introduce PDE-SSM, a novel architectural block that provides a principled generalization of State Space Models to multi-dimensional spatial data by leveraging learnable PDEs.
    \item We present PDE-SSM\textsubscript{DiT}, a new backbone for generative vision models that utilizes the PDE-SSM for efficient and spatially-aware feature mixing, efficiently implemented via the Fourier transform.
    \item Through extensive experiments on both low- and high-resolution image generation tasks, we show that PDE-SSM\textsubscript{DiT} is an effective generative model, achieving competitive or superior performance to the standard DiT, while being more computationally efficient, especially at higher resolutions.
\end{enumerate}

\begin{figure*}[t]
    \centering
\includegraphics[width=0.99\linewidth]{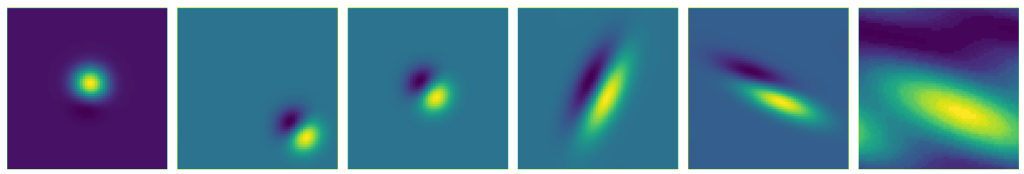}
\vspace{-0.5em}
    \caption{\textbf{Visualizing the PDE-SSM Convolutional Kernels.} By sampling the learnable parameters $\xi = (\mathcal{B}_{\gamma}, \zeta)$, our PDE-SSM can represent a diverse family of convolutional kernels. The examples show kernels that are (from left to right): localized, directionally blurred (anisotropic diffusion), shifted (convection), and a combination of effects. This flexibility allows our PDE-SSM model to learn a rich basis for spatial feature mixing, including non-local connections.}
    \label{fig:convs}
    \vspace{-0.75em}
\end{figure*}

\section{From 1D SSM to a Spatial PDE-SSM}
\label{sec:pde_ssm}

In this section, we develop the PDE-SSM as a principled generalization of State Space Models from 1D sequences to multi-dimensional spatial data. We begin by reformulating the 1D SSM from a differential operator perspective in \Cref{sec:diff_view_ssm}, which provides a natural bridge to our spatial, 2D formulation in \Cref{sec:spatial_ssm}, followed by its efficient implementation in \Cref{sec:computingPDESSM}.

\subsection{A Differential Operator View of SSMs}
\label{sec:diff_view_ssm}

We start by utilizing the known connection between SSMs and ODEs, and their solutions \citep{evans2022partial}: a linear time-invariant SSM (\Cref{eq:ssm}), can be rewritten as the solution to an ODE:
\begin{equation}
    \mathcal{L}h(t) = Bu(t), \quad \text{where} \quad \mathcal{L} = \frac{d}{dt} - A.
\end{equation}
Here, $u(t)$ is the input signal, $h(t)$ is the hidden state at time $t$, and $A, \ B$ are learnable matrices. $\mathcal{L}$ is a first-order linear differential operator that acts on the hidden state $h(t)$. The solution can be expressed as a convolution with the Green's function (or impulse response) of the operator $\mathcal{L}$:
\begin{equation}
    \label{eq:ssm_intform}
    h(t) = (\mathcal{G}_A \star Bu)(t), \quad \text{where} \quad \mathcal{G}_A(t) = e^{tA} \mathbb{I}(t \ge 0).
\end{equation}
The computation described in \Cref{eq:ssm_intform} is \emph{convolutional} --- which is key to the efficiency of modern SSMs \citep{gu2021efficiently}. By structuring the matrix $A$, the convolutional kernel $\mathcal{G}_A$ can be computed efficiently, 
in some cases in the Fourier domain \citep{gu2021efficiently}, 
allowing the model to capture long-range dependencies in near-linear time.

\begin{figure*}[t] 
  \centering
  \begin{AlgWrap}
    \captionof{algorithm}{\textbf{PDE-SSM Forward Pass}}
    \label{alg:pde-ssm-coupled}

    \begin{algorithmic}[1]
    \makeatletter
\setcounter{ALG@line}{0}
\makeatother
      \Require Input tensor $\mathbf{u} \in \mathbb{R}^{B \times C_{\rm in} \times H \times W }$.
      \Require Learnable parameters $\mathbf{R} \in \mathbb{R}^{C_{\rm hid} \times C_{\rm in}}, \mathbf{\Gamma}_0, \mathbf{\Gamma}_x, \mathbf{\Gamma}_y, \mathbf{K}_{xx}, \dots,
      \in \mathbb{R}^{C_{\rm hid} \times C_{\rm hid}}, \tau$.
      \Statex

      \State \AlgSection{1. Input Embedding in Fourier Domain}
      \State $\mathbf{u}' \leftarrow \mathrm{Conv}_{1\times1}(\mathbf{u}, \mathbf{R})$ \Comment{Project to $C_{\rm hid}$ channels}
      \State $\hat{\mathbf{u}}' \leftarrow \mathrm{FFT}(\mathbf{u}')$ \Comment{Transform to frequency domain}
      \State Generate $\mathbf{k} = (k_x, k_y)$ as the discrete frequency grid.
      \State $\hat{\mathcal{B}}(\mathbf{k}) \leftarrow \mathbf{\Gamma}_0 + i k_x \mathbf{\Gamma}_x + i k_y \mathbf{\Gamma}_y$ \Comment{Matrix symbol for $\mathcal{B}$}
      \State $\hat{\mathbf{v}}(\mathbf{k}) \leftarrow \hat{\mathcal{B}}(\mathbf{k}) \cdot \hat{\mathbf{u}}'(\mathbf{k})$ \Comment{Apply embedding (matrix–vector product)}
      \Statex

      \State \AlgSection{2. Coupled PDE Evolution in Fourier Domain}
      \State $\mathbf{\Lambda}(\mathbf{k}) \leftarrow -(\mathbf{k}^\top \mathbf{K} \mathbf{k}) + \mathbf{R}_m + i(\mathbf{k} \cdot \mathbf{B})$ \Comment{Assemble $C_{\rm hid}\!\times\! C_{\rm hid}$ matrix for each $\mathbf{k}$}
      \State $\hat{\mathcal{G}}_{\zeta}(\mathbf{k}) \leftarrow \exp(\tau\, \mathbf{\Lambda}(\mathbf{k}))$ \Comment{Green's function symbol}
      \State $\hat{\mathbf{h}}(\mathbf{k}) \leftarrow \hat{\mathcal{G}}_{\zeta}(\mathbf{k}) \cdot \hat{\mathbf{v}}(\mathbf{k})$ \Comment{Apply PDE operator (channelwise matrix–vector)}
      \Statex

      \State \AlgSection{3. Return to Spatial Domain}
      \State $\mathbf{h} \leftarrow \mathrm{iFFT}(\hat{\mathbf{h}})$ \Comment{Back to spatial domain}
      \State \Return $\mathbf{h}$
    \end{algorithmic}
  \end{AlgWrap}
\end{figure*}

\subsection{Generalizing to Space: The PDE-SSM Formulation}
\label{sec:spatial_ssm}

To extend the SSM framework from a 1D sequence, temporal domain to a multi-dimensional spatial domain $\mathbf{x} \in \mathbb{R}^{d}$, we replace the ODE with a PDE. This is a natural generalization, because PDEs are the fundamental mathematical framework for describing systems that evolve over both time and space. The resulting operator, which we call \emph{PDE-SSM}, maps an input feature map $u(\mathbf{x})$ to an output feature map $h(\mathbf{x})$ via a spatial convolution, as follows:
\begin{equation}
    \label{eq:ssmnd_intform}
    h(\mathbf{x}) =  (\mathcal{G}_{\zeta} \star \mathcal{B}_{\gamma} u)(\mathbf{x}).
\end{equation}

Both $B_\gamma$ and $G_\zeta$ are linear spatial operators and could theoretically be combined into a single operation. In practice, we apply them sequentially in the Fourier domain, since we compute the action of the matrix exponential through matrix--vector products rather than forming it explicitly.

The PDE-SSM operation, described in \Cref{eq:ssmnd_intform}, consists of two stages: (i) an input embedding $\mathcal{B}_{\gamma}u$ to obtain the initial state; and (ii) a spatial evolution defined by the Green's function $\mathcal{G}_{\zeta}$ of our learnable PDE. In what follows, we formalize these two operations.

\paragraph{The Embedding Operator $\mathcal{B_{\gamma}}$.}
The operator $\mathcal{B}_{\gamma}$ is a learnable embedding layer that transforms the input $u$  into the hidden space. For a 2D image, which is at the focus of this paper, it reads:
\begin{equation}
    \label{eq:B}
    \mathcal{B}_{\gamma} u = (\gamma_0 I + \gamma_x \frac{\partial}{\partial x} + \gamma_y \frac{\partial}{\partial y}) (R \star u)(\mathbf{x}).
\end{equation}
Here, $R$ is a $1 \times 1$ (also known as pointwise) 2D convolution that mixes the input channels of $u$. The learnable scalars $\gamma_0, \gamma_x, \gamma_y$ then create a new feature map from a weighted sum of the channel-mixed features and their spatial gradients $\frac{\partial}{\partial x}$ and $\frac{\partial}{\partial y}$.

\paragraph{The PDE Evolution and its Green's Function $\mathcal{G}_{\zeta}$.}
The core of our model is the diffusion-convection-reaction PDE from \Cref{eq:ssm_nd}, which governs the evolution of the state $h(\mathbf{x}, t)$ from an initial condition $h(\mathbf{x}, t=0) = \mathcal{B}_{\gamma}u(\mathbf{x})$. The PDE-SSM state is obtained by solving the PDE \eqref{eq:ssm_nd} in the Fourier domain to derive an analytic expression for its Green's function. To this end, let $\hat{h}(\mathbf{k}, t)$ be the spatial Fourier transform of $h(\mathbf{x}, t)$, where $\mathbf{k} = (k_x, k_y)$ is the corresponding frequency vector. In Fourier space, the spatial derivatives $\frac{\partial}{\partial x}$ and $\frac{\partial}{\partial y}$ transform into multiplication by $i\mathbf{k}$: $\nabla \to i\mathbf{k}$. Overall, applying the Fourier transform to the PDE yields:
\begin{align}
    \frac{\partial \hat{h}(\mathbf{k}, t)}{\partial t} &= (i\mathbf{k})^T K (i\mathbf{k})\hat{h}(\mathbf{k}, t) + \mathbf{b} \cdot (i\mathbf{k})\hat{h}(\mathbf{k}, t) + r\hat{h}(\mathbf{k}, t) \\
    &= \underbrace{(-\mathbf{k}^T K \mathbf{k} + r + i(\mathbf{b} \cdot \mathbf{k}))}_{\lambda(\mathbf{k})} \hat{h}(\mathbf{k}, t).
\end{align}
This transforms the PDE into a simple, linear ODE for each frequency $\mathbf{k}$, with the solution:
\begin{equation}
    \hat{h}(\mathbf{k}, \tau) = e^{\tau \lambda(\mathbf{k})} \hat{h}(\mathbf{k}, 0)
\end{equation}
The term $e^{\tau \lambda(\mathbf{k})}$ is the Fourier transform of the Green's function, $\hat{\mathcal{G}}_{\zeta}(\mathbf{k})$, also known as the symbol of the operator, and the  kernel is defined in the frequency domain by the parameters $\zeta = (K, \mathbf{b}, r, \tau)$:
\begin{equation}
    \label{eq:symbol}
    \hat{\mathcal{G}}_{\zeta}(\mathbf{k}) = \exp\left(\tau(-\mathbf{k}^T K \mathbf{k} + r + i(\mathbf{b} \cdot \mathbf{k}))\right).
\end{equation}

\paragraph{Theoretical Properties of PDE-SSM.} Beyond computational efficiency, the PDE-SSM operator inherits structural guarantees from its PDE formulation. 
The diffusion term can provide structural damping. However, without explicit constraints on $K$, $b$, $r$, or the coupled operators, the reaction and multi-channel coupling terms may introduce eigenvalues with positive real parts. Therefore, unconditional spectral stability would require additional constraints.
Expressivity is preserved through the combination of diffusion, convection, and reaction terms, which together span a rich class of kernels that includes localized filters, global translations, and nonlocal smoothers, as illustrated in \Cref{fig:convs}. Nonlocality is inherent as well: the integration time $\tau$ directly controls the receptive field, and as $\tau$ increases, the operator aggregates information across arbitrarily large spatial scales, paralleling the long-range memory of 1D SSMs but in the spatial domain. These guarantees make PDE-SSM a principled and theoretically grounded alternative to attention for spatial mixing.  

The closed-form expression in Eq.~(10) enables efficient computation while allowing the learnable parameters to encode physically meaningful inductive biases:  
\begin{itemize}[leftmargin=2em]
    \item \textbf{Diffusion ($K$):} The term $-\tau \mathbf{k}^\top K \mathbf{k}$ is a non-positive quadratic form, acting as a low-pass filter that damps high frequencies. Anisotropic $K$ induces direction-dependent smoothing.  
    \item \textbf{Convection ($\mathbf{b}$):} The term $i\tau(\mathbf{b}\cdot \mathbf{k})$ induces a phase shift $e^{i\tau(\mathbf{b}\cdot \mathbf{k})}$, corresponding to a spatial translation via the Fourier shift theorem, enabling directed nonlocal relationships.  
    \item \textbf{Reaction ($r$):} The scalar $\tau r$ uniformly scales responses across all frequencies, modeling global amplification ($r>0$) or suppression ($r<0$).  
    \item \textbf{Time ($\tau$):} The integration time modulates the strength of all effects: as $\tau \to 0$, the operator approaches the identity, while larger $\tau$ values yield increasingly global kernels.  
\end{itemize}


The reaction term $r$ is learned rather than fixed. Diffusion damps high-frequency components, while reaction provides uniform amplification or suppression across frequencies. Since $K$, $b$, and $r$ are optimized jointly, their balance is determined by the flow-matching objective.

We illustrate different kernels obtained by our PDE-SSM convolution kernel in \Cref{eq:symbol} in \Cref{fig:convs}. We note that, in the above discussion, we have presented the computation for a single channel. In practice, one needs to consider the multi-channel case. Let $C$ be the number of channels. In this case, in 2D, the matrix $K$ becomes a 4D tensor of size $C \times C \times 2 \times 2$, the vector $\bf b$ becomes a $C \times C \times 2 \times 1$  tensor and $r$ becomes a 
$C \times C \times 1 \times 1$ tensor.
The convolution is performed for each channel pair, and the resulting outputs are aggregated across channels, similar to the way that standard multi-channel 2D convolutions are implemented in modern deep learning frameworks.

\subsection{Efficient Implementation with Multi-Channel Coupling}
\label{sec:computingPDESSM}

The analytic form of the kernel in the Fourier domain in \Cref{eq:symbol} allows for an efficient implementation. To model rich interactions where channels interact, i.e., mix, we generalize the PDE-SSM parameters $\zeta = (K, \mathbf{b}, r, \tau)$ from scalars to matrices that operate on the channel dimension. This extension transforms the decoupled $C$ PDE-SSM systems into a system of coupled PDEs, which can be solved simultaneously in the Fourier domain. The procedure is detailed in Algorithm~\ref{alg:pde-ssm-coupled}. 
In this formulation, the parameters for embedding ($\mathbf{\Gamma}$'s) and evolution ($\mathbf{K}, \mathbf{B}, \mathbf{R}_m$) are now $C_{\rm hid} \times C_{\rm hid}$ matrices. Consequently, the operations in lines 6, 8, 9, and 10 in \Cref{alg:pde-ssm-coupled} become matrix-matrix operations on the channel direction. 
\paragraph{Computational Complexity.} Channel-coupling impacts the computational cost. While the FFTs remain $\mathcal{O}(C_{\rm hid} \cdot N \log N)$, the Fourier-domain operations have the following complexity:
\textit{Symbol Computation (Lines 5, 8, \Cref{alg:pde-ssm-coupled})}—assembling the $\mathbf{\Lambda}(\mathbf{k})$ matrix for each of the $N$ frequencies involves matrix additions and scalar–matrix products, costing $\mathcal{O}(N \cdot C_{\rm hid}^2)$; \textit{Matrix–Vector Products (Lines 6, 10, \Cref{alg:pde-ssm-coupled})}—these cost $\mathcal{O}(N \cdot C_{\rm hid}^2)$. 
Computing the matrix exponential explicitly at each frequency would incur an additional
$\mathcal{O}(N C_{\rm hid}^3)$ cost. For small channel dimensions, we compute the
Green's function explicitly. For larger $C_{\rm hid}$, we instead compute the action of
the matrix exponential using an Arnoldi approximation with a fixed Krylov dimension of
16, reducing this cost to $\mathcal{O}(N C_{\rm hid}^2)$.
The total complexity is therefore dominated by the Fourier-domain channel operations, leading to an overall cost of $\mathcal{O}(C_{\rm hid} \cdot N \log N + N \cdot C_{\rm hid}^2)$.
The $N \cdot C_{\rm hid}^2$ term dominates when the number of hidden channels is large. 
While both PDE-SSM and standard self-attention share this $\mathcal{O}(N \cdot C_{\rm hid}^2)$ channel-mixing cost, our approach achieves substantial practical speedups by replacing attention's prohibitive $\mathcal{O}(N^2)$ spatial bottleneck with efficient $\mathcal{O}(N \log N)$ spectral mixing.
This asymptotic analysis is unchanged for deeper or wider models, as model scale affects the number of layers and $C_{\rm hid}$ but not the structure of the PDE-SSM operator.

\section{Using PDE-SSM within a Diffusion Transformer}

We now discuss the viability of the PDE-SSM block in the context of Diffusion Transformer (DiT), to address the task of image generation \citep{lipman2023flow, song2021scorebased}. 

\paragraph{Background on Flow-Matching.} We focus our attention on flow-matching techniques as discussed in \citep{lipman2023flow}, where the goal is to train a transformer-based network that solves the following optimization problem:
\begin{eqnarray}
    \label{eq:denoise}
    \min_{\theta} {\mathbb E}_{u, z} \int_0^1 \, \|v_{\theta}( tu + (1-t)z, t) - (u-z) \|^2\, dt,
\end{eqnarray}
where $u$ is the original ("clean") image, $z$ is Gaussian noise and $t$ is time. During training, we are given a corrupt, noisy image
\begin{eqnarray}
\label{eq:ut}
u_t = tu + (1-t)z, 
\end{eqnarray}
and the goal is to train a network $v_{\theta}$ that recovers the  velocity $u-z$. This task is closely related to image denoising, because given $v_{\theta}$, one can estimate the clean image $u$ by setting:
\begin{eqnarray}
    \label{eq:im_denosie}
u \approx u_t + (1-t)v_{\theta}. 
\end{eqnarray}
During training, one uses the clean images by generating randomly corrupted images with different levels of noise, governed by the parameter $t$, and uses the network  $v_{\theta}$ to recover the velocity $u-z$. The loss for a given triplet, $(u,z,t)$, is computed by the integrand of  \eqref{eq:denoise}. The network weights $\theta$ are trained using the AdamW \citep{kingma2017adam,loshchilov2018decoupled} optimizer. 

\textbf{Background on DiTs.} Diffusion Transformer networks use attention-based blocks in the velocity network $v_{\theta}$.
To this end, the image is "patchified", i.e., it is separated into patches, where each patch is comprised of $k\times k$ pixels. Given an image of shape $C \times N \times N$ where $C$ is the number of channels and $N$ is the number of pixels in each axis, each patch is flattened to a $C \cdot k \cdot k$ vector, which is considered as the feature vector that is associated with the patch.
DiTs uses a fully-connected layer on the patches, yielding a coarse image of $N_{patch}^2$ patches with $C_{hid}$ channels, where $N_{patch} = N/k$ (assuming that $N_{patch}$ is an integer). The image is then flattened  to generate a 2D tensor 
with dimensions $C_{hid} \times N_{patch}^2$.
An attention layer is then used between the patches.
Overall, a standard DiT block takes the computational form of:
\begin{eqnarray}
\label{eq:DiTblock}
h_{j+\tfrac{1}{2}} 
&=& h_j + \textsc{Att}\!\left(h_j\right), \\[2pt]
h_{j+1} 
&=& h_{j+\tfrac{1}{2}} + \textsc{MLP}\!\left(h_{j+\tfrac{1}{2}}\right),
\end{eqnarray}
where $\textsc{Att}$ is an attention layer \citep{vaswani2017attention}, and $\textsc{MLP}$ is a standard multilayer-perceptron. Note that the MLP operates on all patches individually (similar to a dilated convolution), while the attention models generate interactions between patches.
The PDE-SSM operator itself is linear. Nonlinearities in the full architecture arise from the pointwise MLPs interleaved with PDE-SSM inside each DiT block.

\paragraph{Patch size and complexity.} The patch size $k$ in DiTs critically controls the computation. Large $k$ inflates the MLP—whose dense layers scale as $C_{\mathrm{hid}}\!\times\! C\!\times\! k\!\times\! k$—but reduces the number of patches to $N_{\mathrm{patch}}^2 = N^2/k^2$, so attention couples fewer tokens; since attention cost grows quadratically in the number of patches, this yields a scaling proportional to $(N_{\mathrm{patch}}^2 \times N_{\mathrm{patch}}^2)=N_{\mathrm{patch}}^4$. Conversely, a small $k$ shrinks the MLP yet explodes the patch count, making attention dominant. Empirically, a smaller $k$ improves generation quality \citep{peebles2023scalable}, further emphasizing attention as the bottleneck. Our PDE-SSM variant keeps the DiT block of \eqref{eq:DiTblock} unchanged except for swapping $\textsc{Att}$ with a PDE-SSM block. By the analysis in \Cref{sec:computingPDESSM}, this yields a complexity of $\mathcal{O}\!\left(N_{\mathrm{patch}}^2 \log N_{\mathrm{patch}}\right)$, allowing us to choose small $k$ without incurring prohibitive attention cost.

\section{Experiments}
\label{sec:experiments}

\begin{figure*}[t]
    \centering
    \begin{subfigure}[t]{0.28\linewidth}
        \includegraphics[width=\linewidth]{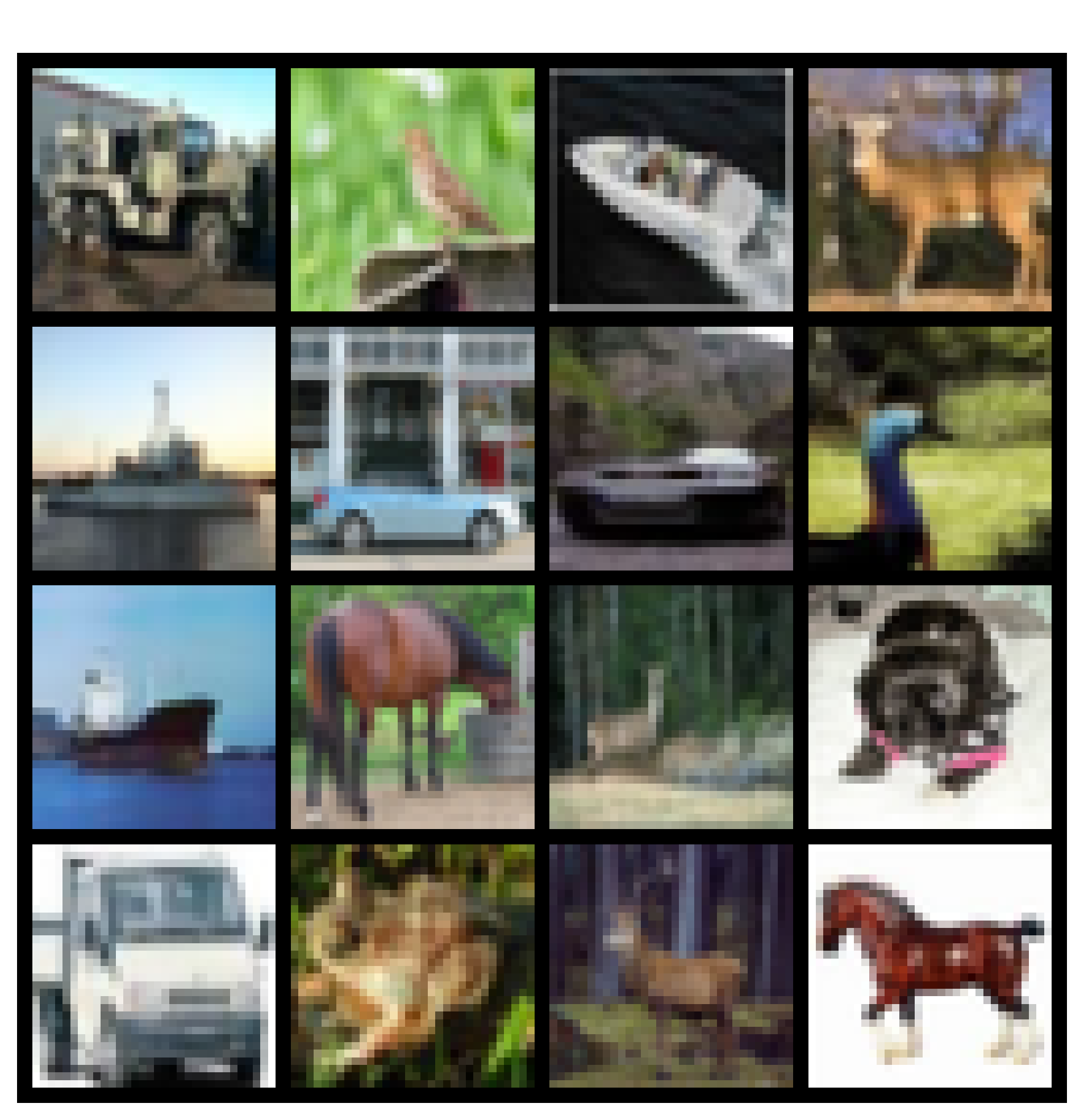}
        \caption{Real images}
        \label{fig:cifar-real}
    \end{subfigure}\hfill
    \begin{subfigure}[t]{0.28\linewidth}
        \includegraphics[width=\linewidth]{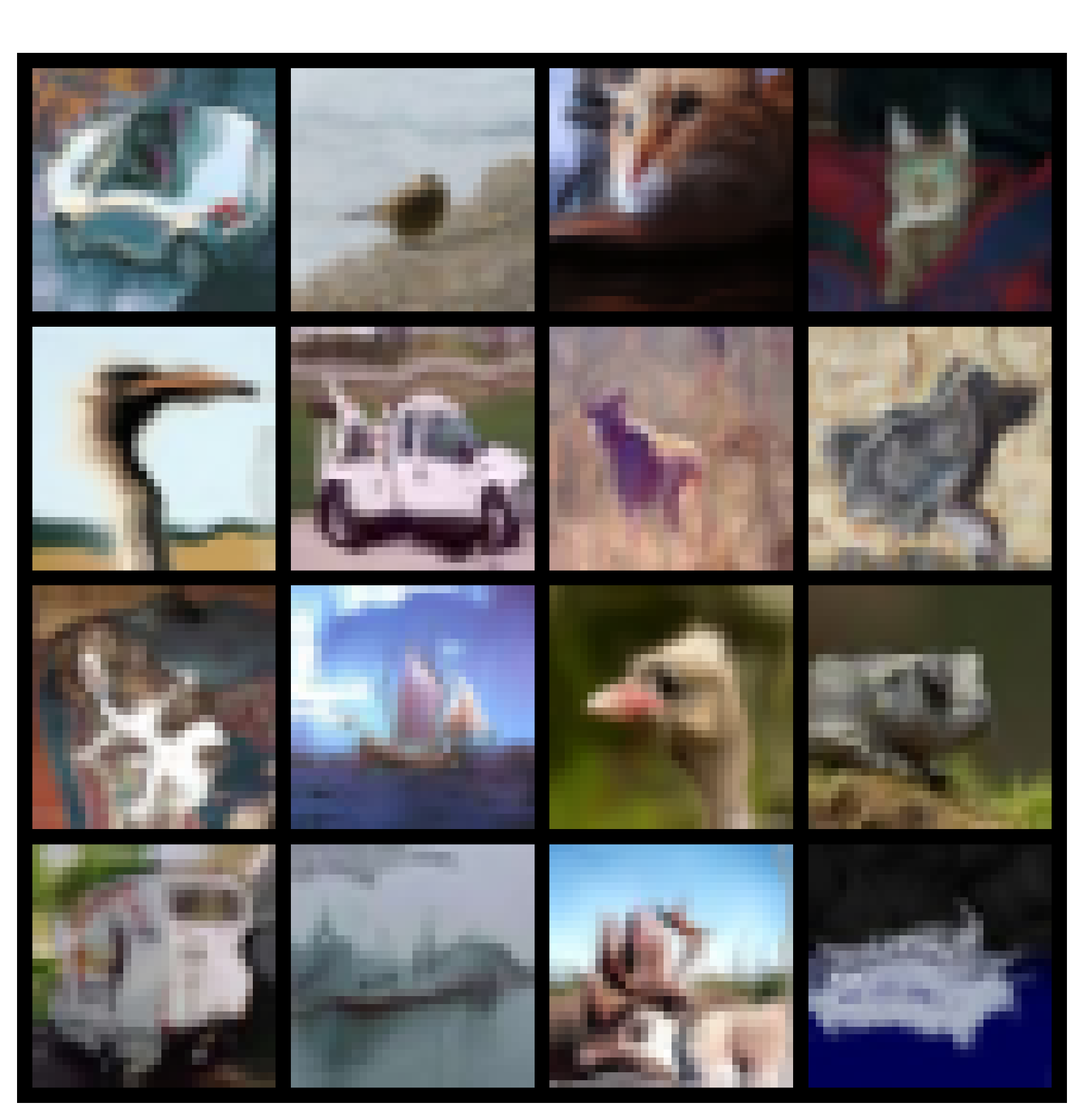}
        \caption{DiT}
        \label{fig:cifar-attn}
    \end{subfigure}\hfill
    \begin{subfigure}[t]{0.28\linewidth}
        \includegraphics[width=\linewidth]{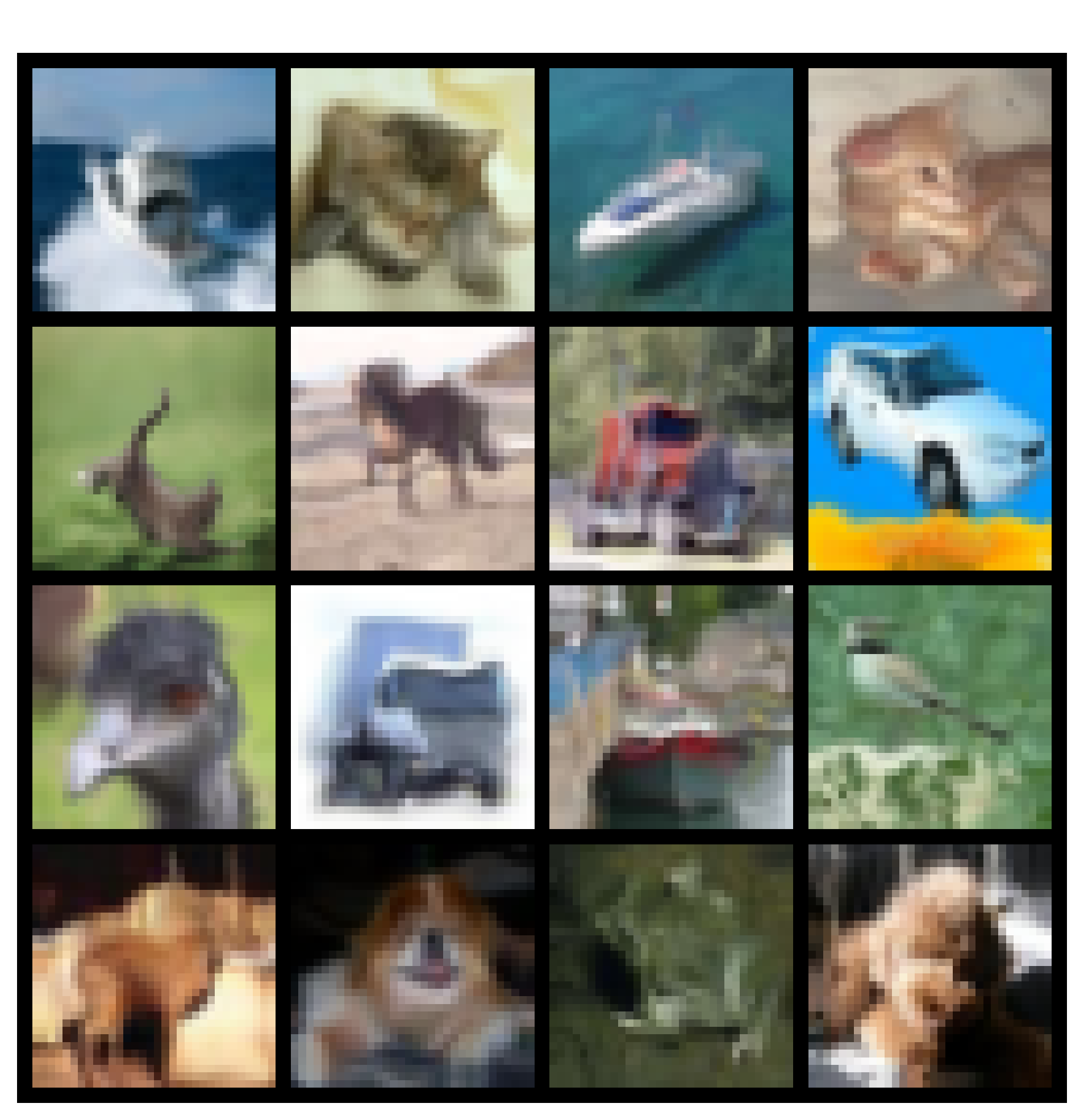}
        \caption{PDE-SSM-DiT}
        \label{fig:cifar-pdessm}
    \end{subfigure}
    \caption{CIFAR-10 Images: \textbf{(a)} real images; \textbf{(b)} DiT; \textbf{(c)} PDE-SSM-DiT. Visual quality is comparable, in congruence with Table~\ref{tab:cifar10}.}
    \label{fig:CIFAR10}
\end{figure*}

We evaluate \emph{PDE-SSM-DiT} by replacing DiT attention with a spatiotemporal state-space module, holding all other components, training schedules, and sampling fixed for a clean ablation. We state the guiding research questions and evaluation protocol, then report results on four benchmarks. We analyze training dynamics via a lightweight FID proxy and study scalability under matched compute as model size and resolution increase. All experiments ran on an NVIDIA RTX6000 Ada GPU. Datasets and hyperparameters are in \Cref{sec:datasets_experimental_settings}.

\paragraph{Research Questions.} Our experiments seek to address the following:
\begin{itemize}[leftmargin=2em]
  \item \textbf{RQ1} (\emph{PDE-SSM vs.\ Attention}):
  At fixed training objective, schedule, and sampler, does PDE-SSM-DiT match or improve the generative quality of attention-based DiT?

  \item \textbf{RQ2} (\emph{Plug-and-play replacement}):
  When swapping only the block type (attention $\rightarrow$ PDE-SSM) in an off-the-shelf DiT, do we retain performance without retuning hyperparameters?

  \item \textbf{RQ3} (\emph{Scaling and efficiency}):
  How do wall-clock and asymptotic complexity evolve with image and patch resolution compared to attention?

  \item \textbf{RQ4} (\emph{Domain robustness}):
  Are the gains consistent across class-diverse (CIFAR-10, ImageNet) and structure- or texture-centric datasets (Oxford-Flowers, LSUN-Churches)?
\end{itemize}

\begin{figure*}[t]
    \centering
    \begin{subfigure}[t]{0.28\linewidth}
        \includegraphics[width=\linewidth]{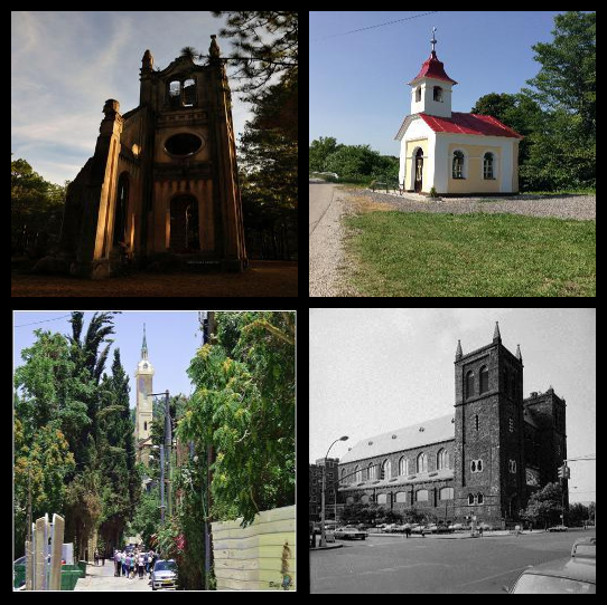}
        \caption{Real images}
        \label{fig:churches-real}
    \end{subfigure}\hfill
    \begin{subfigure}[t]{0.28\linewidth} 
        \includegraphics[width=\linewidth]{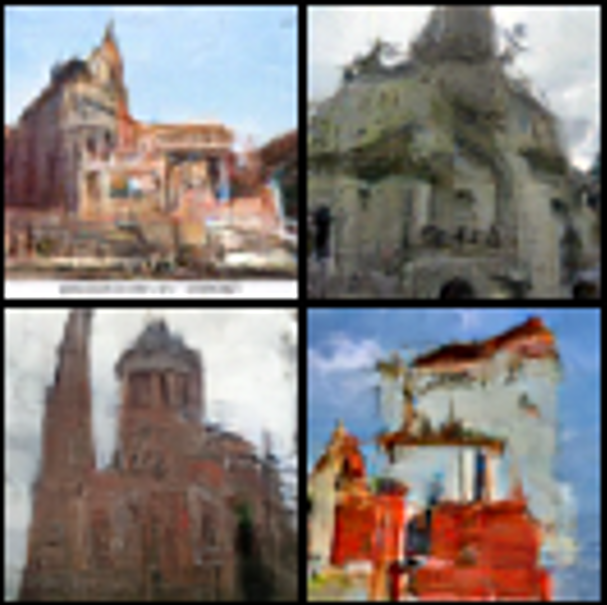}
        \caption{DiT}
        \label{fig:churches-attn}
    \end{subfigure}\hfill
    \begin{subfigure}[t]{0.28\linewidth}
        \includegraphics[width=\linewidth]{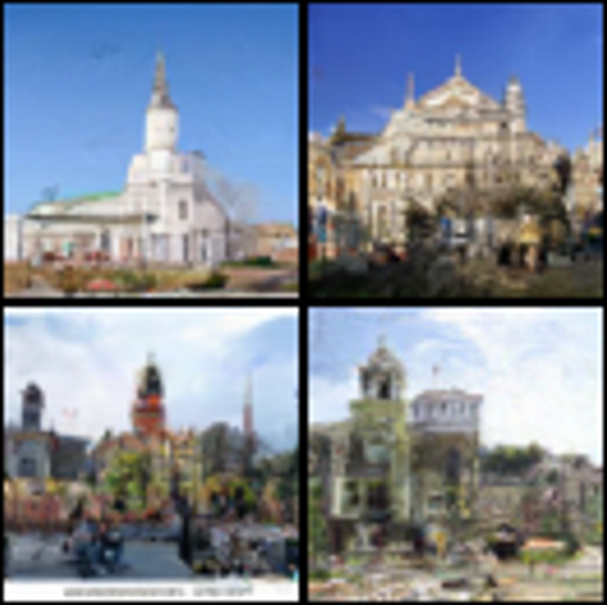}
        \caption{PDE-SSM-DiT}
        \label{fig:churches-pdessm}
    \end{subfigure}
    \caption{LSUN-Churches generations: \textbf{(a)} real images; \textbf{(b)} DiT; \textbf{(c)} PDE-SSM-DiT.}
    \label{fig:churches}
\end{figure*}

\paragraph{Evaluation Protocol.}
We conduct all experiments in the official DiT codebase (\url{https://github.com/facebookresearch/DiT}).
We only replace the attention block from \eqref{eq:DiTblock} with our PDE-SSM block defined in \eqref{eq:ssm_nd}.
All remaining components are held fixed, including normalization layers, positional embeddings, loss definitions, training schedules, time discretization, and the sampling procedure.
We report results for three systems: DiT (unmodified DiT with attention), UNet (a standard diffusion baseline), and PDE-SSM-DiT (DiT where attention is replaced by PDE-SSM).
Because end-to-end diffusion and flow-matching pipelines vary across prior work in the training domain (pixel vs.\ latent) \citep{rombach2022high}, time grids, learning objectives \citep{tong2023improving, song2023consistency}, guidance mechanisms \citep{ho2022classifier, dhariwal2021diffusion}, and samplers \citep{lu2022dpm}, our protocol yields a direct, plug-and-play comparison that isolates the architectural swap under matched settings.
Together with the results reported below, this controlled evaluation highlights the effectiveness of PDE-SSM while minimizing confounding changes in training or inference.

\paragraph{Training and sampling.}
Unless stated otherwise, we use the default DiT hyperparameters and the standard sampling integrators from the public release cited above.
Although these settings were tuned for the original DiT, we keep optimization schedules and data augmentations unchanged so that differences isolate the effect of the block replacement, directly addressing \textbf{RQ2}.
We report mean metrics over three random seeds when applicable.

\paragraph{Metrics.} We report Fréchet Inception Distance (FID)
and, where relevant, mean-squared-error (MSE).

\paragraph{Datasets.} We utilize the CIFAR-10 \citep{krizhevsky2009learning}, CelebA-HQ \citep{karras2017progressive}, ImageNet$64$ \citep{deng2009imagenet}, Imagenet$256$ \citep{Krizhevsky2012Imagenet}, LSUN-Churches \citep{yu2015lsun}, and Oxford-Flowers \citep{nilsback2008automated}. CIFAR-10 and ImageNet stress class diversity; CelebA-HQ, Oxford-Flowers, and LSUN-Churches emphasize fine structure and global coherence, addressing \textbf{RQ4}. We provide additional details regarding the datasets and experimental settings in Appendix \ref{sec:datasets_experimental_settings}.

\subsection{Image Generation with \texorpdfstring{PDE-SSM-DiT}{PDE-SSM-DiT}}
\label{subsec:image_gen}

\begin{wraptable}{r}{0.55\linewidth}
\vspace{-1em}
\centering
\caption{{PDE-SSM-DiT} matches attention-based DiT and U-Net performance on CIFAR-10, indicating \emph{quality parity at equal settings}, isolating the effect of the block swap (\textbf{RQ1, RQ2}).}
\begingroup
\setlength{\tabcolsep}{1pt} 
\renewcommand{\arraystretch}{1} 
\footnotesize
\begin{tabular}{lccc}
\toprule
{Method} & {\#Params (M)} & {MSE$\downarrow$} & {FID (50K)$\downarrow$} \\
\midrule
DiT & 29.6 & $3.82 \times 10^{-2}$ & 4.25 \\
U-Net & 32.6 & $3.78 \times 10^{-2}$ & 4.19 \\
{PDE-SSM-DiT} (Ours) & 34.2 & $\mathbf{3.76 \times 10^{-2}}$ & \textbf{4.18} \\
\bottomrule
\end{tabular}

\label{tab:cifar10}
\endgroup
\vspace{-1.5em}
\end{wraptable}

\paragraph{CIFAR-10.}
CIFAR-10 is a compact and diverse benchmark. We compare a small DiT (\(\sim\)30--34M parameters), a U-Net of comparable size, and {PDE-SSM-DiT} with a matched parameter budget. Results in \Cref{tab:cifar10} shows parity in fidelity, while qualitative samples in \Cref{fig:CIFAR10} further confirm that the block swap preserves generative quality while unlocking the scaling benefits analyzed in \Cref{subsec:efficiency} (\textbf{RQ1, RQ2}).

\begin{wraptable}{r}{0.42\textwidth}
\vspace{-1em}
\footnotesize
\centering
\caption{Unconditional Generation on ImageNet-$64{\times}64$.}
\label{tab:imagenet64_uncond_fid}
\begin{tabular}{lcc}
\toprule
{Method} & {Params (M)} & {FID$\downarrow$} \\
\midrule
DiT & 33 & 22.9 \\
U-Net & 32 & 22.2 \\
{PDE-SSM-DiT} (Ours) & \textbf{31} & \textbf{22.1} \\
\bottomrule
\end{tabular}
\vspace{-1em}
\end{wraptable}

\paragraph{ImageNet.} 

On ImageNet$64$, {PDE-SSM-DiT} achieves a modest yet consistent improvement over DiT under identical training and sampling conditions, as reported in \Cref{tab:imagenet64_uncond_fid}. The model also remains competitive with U-Net at comparable parameter counts, while being strictly plug-and-play (\textbf{RQ1, RQ2, RQ4}).
Literature-reported results are provided for reference, though they may differ in experimental setup and sample counts. 
We present additional results for \textbf{ImageNet} in \Cref{app:imagenet}.


\begin{figure*}[t]
    \centering
    \begin{subfigure}[t]{0.28\linewidth}
        \includegraphics[width=\linewidth]{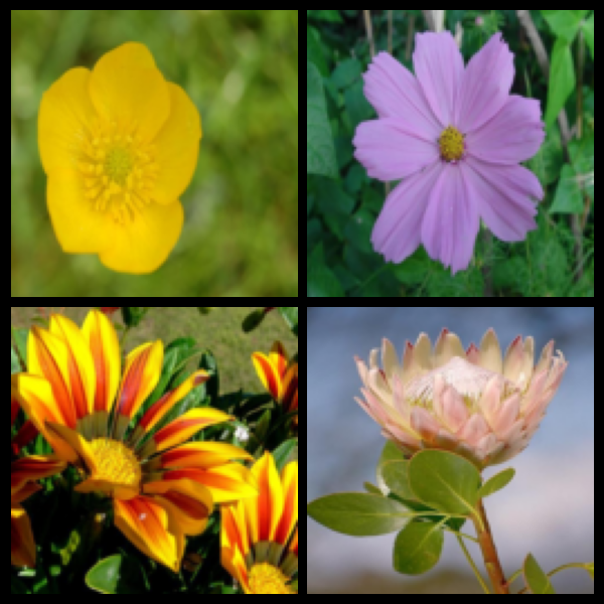}
        \caption{Real images}
        \label{fig:flowers-real}
    \end{subfigure}\hfill
    \begin{subfigure}[t]{0.28\linewidth}
        \includegraphics[width=\linewidth]{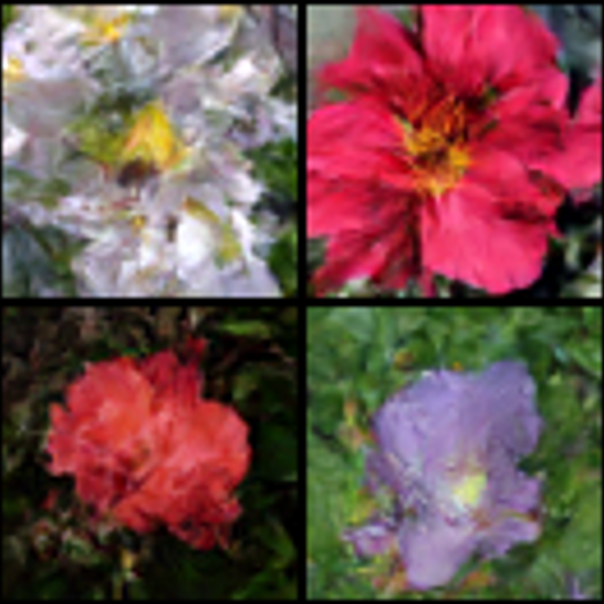}
        \caption{DiT}
        \label{fig:flowers-attn}
    \end{subfigure}\hfill
    \begin{subfigure}[t]{0.28\linewidth}
        \includegraphics[width=\linewidth]{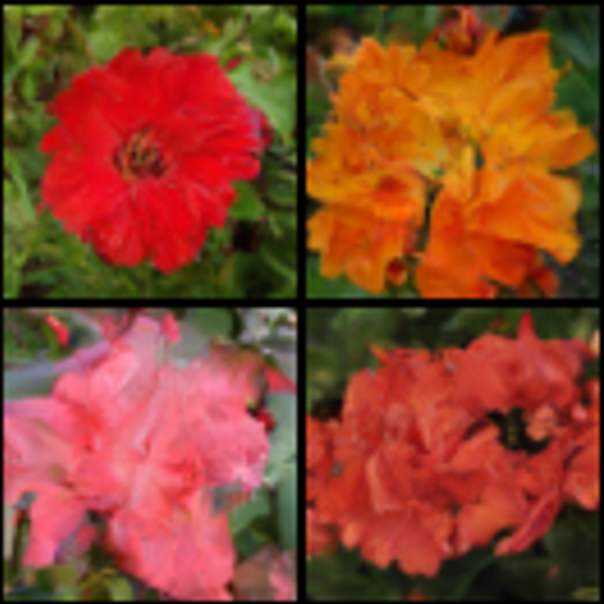}
        \caption{PDE-SSM-DiT}
        \label{fig:flowers-pdessm}
    \end{subfigure}
    \caption{Oxford-Flowers generations: \textbf{(a)} real images; \textbf{(b)} DiT; \textbf{(c)} PDE-SSM-DiT.}
    \label{fig:flowers}
\end{figure*}

\begin{wraptable}{r}{0.45\textwidth}
\vspace{-1.5em}
\centering
\caption{FID (2048 features) on CelebA-HQ$64$, LSUN-Churches and Oxford-Flowers.}
\label{tab:church_flowers_fid}
\begingroup
\setlength{\tabcolsep}{2pt}
\footnotesize
\begin{tabular}{@{}lccc@{}}
\toprule
\multirow{2}{*}{Method} & CelebA & LSUN & Oxford \\
 & HQ$64$ & Churches & Flowers \\
\midrule
DiT                  & 9.17  & 55.04 & 61.50 \\
U-Net                & 11.06 & 51.56 & 57.20 \\
{PDE-SSM-DiT} (Ours) & \textbf{7.01} & \textbf{40.71} & \textbf{49.56} \\
\bottomrule
\end{tabular}
\endgroup
\vspace{-1em}
\end{wraptable}

\paragraph{CelebA-HQ$64$, CelebA-HQ$256$, LSUN-Churches} and \textbf{Oxford-Flowers:}
On structure- and texture-centric datasets, {PDE-SSM-DiT} improves FID relative to DiT, indicating benefits for long-range coherence with small patches (\textbf{RQ1, RQ4}). Samples are shown in  \Cref{fig:churches}, \Cref{fig:flowers} and \Cref{app:celcahq_samples}, while results are reported in \Cref{tab:church_flowers_fid,tab:CelebAHQ256_FID}. 
\subsection{Computational Efficiency and Scaling}
\label{subsec:efficiency}

\begin{table}[t]
\caption{Training results on CelebA HQ 256}
\centering
\begin{tabular}{lccccc}
\toprule
Architecture & Parameters & FID & MSE\\
\midrule
DiT & 32.99 M & 21.08 & 8.02e-2\\
PDE-SSM-DiT(Ours) & 26.48 M & 18.36 & 5.37e-2\\
\bottomrule
\end{tabular}

\label{tab:CelebAHQ256_FID}
\end{table}

\begin{table}[t]
\centering
\footnotesize
\caption{DiT vs.\ {PDE-SSM-DiT} training step runtimes (in seconds). Parameters in millions (M).}
\label{tab:cost}
\small
\setlength{\tabcolsep}{3pt} 
\begin{tabular}{cccccc}
\toprule
{Image} & {Patch} & \multicolumn{2}{c}{{Params (M)}} & \multicolumn{2}{c}{{Time (s)}} \\
\cmidrule(lr){3-4}\cmidrule(lr){5-6}
 & {Size}  & {Attn} & {PDE-SSM} & {Attn} & {PDE-SSM} \\
\midrule
32  & 2  & 9.01 & 12.76 & $1.09{\times}10^{-1}$ & $2.14{\times}10^{-2}$ \\
32  & 4  & 8.98 & 12.76 & $3.34{\times}10^{-2}$ & $2.40{\times}10^{-2}$ \\
32  & 8  & 9.13 & 12.76 & $1.60{\times}10^{-2}$ & $2.40{\times}10^{-2}$ \\
\midrule
64  & 2  & 9.31 & 12.78 & $4.66{\times}10^{-1}$ & $5.58{\times}10^{-2}$ \\
64  & 4  & 9.06 & 12.78 & $9.92{\times}10^{-2}$ & $5.77{\times}10^{-2}$ \\
64  & 8  & 9.15 & 12.78 & $3.24{\times}10^{-2}$ & $5.84{\times}10^{-2}$ \\
\midrule
128 & 2  & 10.49 & 12.88 & $3.00{\times}10^{0}$  & $2.24{\times}10^{-1}$ \\
128 & 4  & 9.35  & 12.88 & $4.70{\times}10^{-1}$ & $2.05{\times}10^{-1}$ \\
128 & 8  & 9.22  & 12.88 & $1.44{\times}10^{-1}$ & $1.96{\times}10^{-1}$ \\
\midrule
256 & 2  & 15.21 & 13.27 & $3.42{\times}10^{1}$  & $8.48{\times}10^{-1}$ \\
256 & 4  & 10.53 & 13.27 & $3.32{\times}10^{0}$  & $8.47{\times}10^{-1}$ \\
256 & 8  & 9.52  & 13.27 & $5.07{\times}10^{-1}$ & $8.96{\times}10^{-1}$ \\
\bottomrule
\end{tabular}
\end{table}

A key benefit of {PDE-SSM-DiT} is favorable scaling with the number of patches. While attention scales as \(O(N_{\text{patch}}^{4})\) when accounting for 2D tokenization, our implementation scales as \(O(N_{\text{patch}}^{2}\log N_{\text{patch}})\). In practice, attention encourages large patch sizes \(k{\times}k\) to curb quadratic attention cost, shrinking the receptive field for a fixed token budget. {PDE-SSM-DiT} maintains efficiency at small patches, preserving long-range dependencies without the attention bottleneck (\textbf{RQ3}).
Table~\ref{tab:cost} quantifies parameter counts and per-step wall-clock across resolutions, and we show additional timing results in \Cref{sec:timig}. 
In \Cref{sec:ssm_maps}, we show the spatial effect of the SSM model.

\noindent\textbf{Main Takeaways.} (i) At small patches and higher resolutions, {PDE-SSM-DiT} yields substantial wall-clock savings vs.\ attention, enabling finer tokenization without exploding cost. (ii) At very large patches, attention narrows the gap, but this comes at the expense of the global receptive field.



\section{Conclusion and Discussion}

\paragraph{Limitations.}
PDE-SSM is a structured spatial mixing operator and differs from self-attention in an important way. Self-attention computes content-dependent pairwise interactions between tokens, which can be beneficial when precise instance-specific correspondence between distant regions is required. In contrast, PDE-SSM imposes a spectral, translation-equivariant mixing structure whose adaptivity comes through learned channels, nonlinear layers, and depth. This makes the method attractive when global spatial mixing and favorable scaling are the main bottlenecks, but it may be less suitable for tasks requiring explicit token-to-token matching or highly instance-specific routing. We discuss broader impact of out method in \Cref{app:broader_impact}.

\paragraph{Conclusion.} We extend state-space models from 1D to $n^d$ and use the resulting PDE-SSM as a drop-in replacement for attention inside DiT for image generation. The key idea is to replace the 1D ODE with a space-time PDE that combines diffusion, advection, and reaction with a trainable integration horizon, producing learned nonlocal kernels whose field of view ranges from a few pixels to the entire image. Empirically, PDE-SSM-DiT matches or improves upon attention-based DiT while scaling more favorably: 
attention has a spatial complexity of $O(N^2)$ in tokens $N$, whereas our spectral solver reduces this spatial mixing to $O(N \log N)$, 
enabling faster training and more favorable scaling at high resolutions.

A broader takeaway is that strong generators share nonlocal mixing. UNets achieve it through multiscale downsampling, attention through pairwise correlations, and SSM or PDE-SSM through dynamical evolution that couples distant states. The emerging principle is simple: nonlocality is all you need.


\section*{Acknowledgments and Disclosures}
 ME acknowledges support from the the Israel Science Foundation through a personal grant (ISF 3062/26) and an
equipment grant (ISF 3066/26), as well as support from the Israeli Ministry of Innovation, Science \& Technology.
The authors declare no competing interests.

\bibliography{iclr2025_conference}

@article{yan2023diffusion,
  title        = {Diffusion Models Without Attention},
  author       = {Yan, Jing Nathan and Gu, Jiatao and Rush, Alexander M.},
  journal      = {Computer Vision and Pattern Recognition},
  year         = {2023},
}

@article{ceni2025message,
  title={Message-Passing State-Space Models: Improving Graph Learning with Modern Sequence Modeling},
  author={Ceni, Andrea and Gravina, Alessio and Gallicchio, Claudio and Bacciu, Davide and Schonlieb, Carola-Bibiane and Eliasof, Moshe},
  journal={arXiv preprint arXiv:2505.18728},
  year={2025}
}

@article{Teng2024DiM,
  title        = {DiM: Diffusion Mamba for Efficient High-Resolution Image Synthesis},
  author       = {Yao Teng and Yue Wu and Han Shi and Xuefei Ning and Guohao Dai and Yu Wang and Zhenguo Li and Xihui Liu},
  journal      = {arXiv.org},
  year         = {2024},
}

@article{Qin2022HighlyAccurateDichotomousImageSegmentation,
  title        = {Highly Accurate Dichotomous Image Segmentation},
  author       = {Xuebin Qin and Hang Dai and Xiaobin Hu and Deng-Ping Fan and Ling Shao and Luc Van Gool},
  journal      = {European Conference on Computer Vision},
  year         = {2022},
}

@inproceedings{agarwal2023spectralsm,
  title        = {Spectral State Space Models},
  author       = {Agarwal, Naman and Suo, Daniel and Chen, Xinyi and Hazan, Elad},
  year         = {2023},
  eprint       = {2312.06837},
  archivePrefix= {arXiv},
  primaryClass = {cs.LG},
  note         = {arXiv:2312.06837 [cs.LG]},
  url          = {https://arxiv.org/abs/2312.06837}
}

@inproceedings{
loshchilov2018decoupled,
title={Decoupled Weight Decay Regularization},
author={Ilya Loshchilov and Frank Hutter},
booktitle={International Conference on Learning Representations},
year={2019},
url={https://openreview.net/forum?id=Bkg6RiCqY7},
}

@inproceedings{fu2023simpleconv,
 title        = {Simple Hardware-Efficient Long Convolutions for Sequence Modeling},
  author       = {Fu, Daniel Y. and Epstein, Elliot L. and Nguyen, Eric and Thomas, Armin W. and Zhang, Michael and Dao, Tri and Rudra, Atri and Ré, Christopher},
  booktitle    = {Proceedings of the 40th International Conference on Machine Learning (ICML 2023)},
  year         = {2023},
  url           = {https://arxiv.org/abs/2302.06646}
}

@inproceedings{goel2022rawaudio,
  title        = {It’s Raw! Audio Generation with State-Space Models},
  author       = {Goel, Karan and Gu, Albert and Donahue, Chris and Ré, Christopher},
  booktitle    = {Proceedings of the 39th International Conference on Machine Learning (ICML 2022)},
  year         = {2022},
  url          = {https://arxiv.org/abs/2202.09729}
}

@inproceedings{shi2023multiresmem,
  title        = {Sequence Modeling with Multiresolution Convolutional Memory},
  author       = {Shi, Jiaxin and Wang, Ke Alexander and Fox, Emily B.},
  booktitle    = {Proceedings of the 40th International Conference on Machine Learning (ICML 2023)},
  year         = {2023},
  eprint       = {2305.01638},
  archivePrefix= {arXiv},
  primaryClass = {cs.LG},
}

@inproceedings{smith2023simplessm,
  title        = {Simplified State Space Layers for Sequence Modeling},
  author       = {Smith, Jimmy T. H. and Warrington, Andrew and Linderman, Scott W.},
  booktitle    = {Proceedings of the International Conference on Learning Representations (ICLR 2023)},
  year         = {2023},
  url          = {https://arxiv.org/abs/2208.04933}
}

@inproceedings{zhang2023discretessm,
  title        = {Effectively Modeling Time Series with Simple Discrete State Spaces},
  author       = {Zhang, Michael and Saab, Khaled K. and Poli, Michael and Dao, Tri and Goel, Karan and Re, Christopher},
  booktitle    = {Proceedings of the 11th International Conference on Learning Representations (ICLR 2023)},
  year         = {2023},
}

@inproceedings{tay2021lra,
  title        = {Long Range Arena: A Benchmark for Efficient Transformers},
  author       = {Yi Tay and Mostafa Dehghani and Samira Abnar and Yikang Shen and Dara Bahri and Philip Pham and Jinfeng Rao and Liu Yang and Sebastian Ruder and Donald Metzler},
  booktitle    = {Proceedings of the 9th International Conference on Learning Representations (ICLR 2021)},
  year         = {2021},
  url          = {https://arxiv.org/abs/2011.04006}
}

@inproceedings{katharopoulos2020transformers,
  title={Transformers are RNNs: Fast Autoregressive Transformers with Linear Attention},
  author={Katharopoulos, Angelos and Vyas, Apoorv and Pappas, Nikolaos and Fleuret, Fran{\c{c}}ois},
  booktitle={International Conference on Machine Learning},
  pages={5156--5165},
  year={2020},
  organization={PMLR}
}

@inproceedings{shen2018efficient,
  title={Efficient Attention: Attention with Linear Complexities},
  author={Shen, Zhuoran and Zhang, Mingyuan and Zhao, Haiyu and Yi, Shuai and Li, Hongsheng},
  booktitle={2018 IEEE Winter Conference on Applications of Computer Vision (WACV)},
  pages={3531--3539},
  year={2018},
  organization={IEEE}
}

@article{song2023consistency,
  title   = {Consistency Models},
  author  = {Yang Song and Prafulla Dhariwal and Mark Chen and Ilya Sutskever},
  journal = {arXiv preprint arXiv:2303.01469},
  year    = {2023},
  url     = {https://arxiv.org/abs/2303.01469}
}

@article{nichol2021improved,
  title   = {Improved Denoising Diffusion Probabilistic Models},
  author  = {Nichol, Alexander and Dhariwal, Prafulla},
  journal = {International Conference on Machine Learning (ICML)},
  pages   = {8162--8171},
  year    = {2021},
  url     = {https://arxiv.org/abs/2102.09672}
}

@article{lu2022dpm,
  title   = {DPM-Solver: A Fast ODE Solver for Diffusion Probabilistic Model Sampling in Around 10 Steps},
  author  = {Lu, Cheng and Zhou, Yuhao and Bao, Fan and Chen, Jianfei and Li, Chongxuan and Zhu, Jun},
  journal = {Advances in Neural Information Processing Systems (NeurIPS)},
  volume  = 35,
  pages   = {5775--5787},
  year    = {2022},
  url     = {https://arxiv.org/abs/2206.00927}
}

@article{ho2022classifier,
  title   = {Classifier-Free Diffusion Guidance},
  author  = {Ho, Jonathan and Salimans, Tim},
  journal = {NeurIPS 2021 Workshop on Deep Generative Models and Downstream Applications},
  year    = {2022},
  url     = {https://arxiv.org/abs/2207.12598}
}

@article{dhariwal2021diffusion,
  title   = {Diffusion Models Beat GANs on Image Synthesis},
  author  = {Dhariwal, Prafulla and Nichol, Alexander},
  journal = {Advances in Neural Information Processing Systems (NeurIPS)},
  volume  = 34,
  pages   = {8780--8794},
  year    = {2021},
  url     = {https://arxiv.org/abs/2105.05233}
}

@article{peebles2023scalable,
  title   = {Scalable Diffusion Models with Transformers},
  author  = {Peebles, William and Xie, Saining},
  journal = {Proceedings of the IEEE/CVF International Conference on Computer Vision (ICCV)},
  year    = {2023},
  url     = {https://arxiv.org/abs/2212.09748}
}

@misc{
lahoti2025chimera,
title={Chimera: State Space Models Beyond Sequences},
author={Aakash Lahoti and Tanya Marwah and Ratish Puduppully and Albert Gu},
year={2025},
url={https://openreview.net/forum?id=Sfmk5amxFu}
}

@article{tong2023improving,
  title   = {Improving and Generalizing Flow-Based Generative Models with minibatch optimal transport},
  author={Tong, Alexander and Fatras, Kilian and Malkin, Nikolay and Huguet, Guillaume and Zhang, Yanlei and Rector-Brooks, Jarrid and Wolf, Guy and Bengio, Yoshua},
  journal = {Transactions on Machine Learning Research},
  year    = {2023},
}

@article{kingma2017adam,
  title   = {Adam: A Method for Stochastic Optimization},
  author  = {Diederik P. Kingma and Jimmy Ba},
  journal = {International Conference on Learning Representations},
  year    = {2014},
}

@book{evans2022partial,
  title={Partial differential equations},
  author={Evans, Lawrence C},
  volume={19},
  year={2010},
  publisher={American mathematical society}
}

@article{song2021scorebased,
  title={Score-Based Generative Modeling through Stochastic Differential Equations},
  author={Song, Yang and Sohl-Dickstein, Jascha and Kingma, Diederik P and Kumar, Abhishek and Ermon, Stefano and Poole, Ben},
  journal={International Conference on Learning Representations (ICLR)},
  year={2021}
}

@article{nilsback2008automated,
  title={Automated flower classification over a large number of classes},
  author={Nilsback, Maria-Elena and Zisserman, Andrew},
  journal={Proceedings of the Indian Conference on Computer Vision, Graphics and Image Processing},
  year={2008},
  pages={722--729}
}

@inproceedings{yu2015lsun,
  title={LSUN: Construction of a large-scale image dataset using deep learning with humans in the loop},
  author       = {Yu, Fisher and Seff, Ari and Zhang, Yinda and Song, Shuran and Funkhouser, Thomas and Xiao, Jianxiong},
  booktitle={Proceedings of the IEEE Conference on Computer Vision and Pattern Recognition (CVPR)},
  year={2015},
  pages={486--494}
}

@article{vaswani2017attention,
  title={Attention is all you need},
  author={Vaswani, Ashish and Shazeer, Noam and Parmar, Niki and Uszkoreit, Jakob and Jones, Llion and Gomez, Aidan N and Kaiser, {\L}ukasz and Polosukhin, Illia},
  journal={Advances in neural information processing systems},
  volume={30},
  year={2017}
}

@article{touvron2023llama,
  title={Llama: Open and efficient foundation language models},
  author       = {Touvron, Hugo and Lavril, Thibaut and Izacard, Gautier and Martinet, Xavier and Lachaux, Marie-Anne and Lacroix, Timothée and Rozière, Baptiste and Goyal, Naman and Hambro, Eric and Azhar, Faisal and Rodriguez, Aurelien and Joulin, Armand and Grave, Edouard and Lample, Guillaume},
  journal={arXiv preprint arXiv:2302.13971},
  year={2023}
}

@inproceedings{liu2021swin,
  title={Swin transformer: Hierarchical vision transformer using shifted windows},
  author={Liu, Ze and Lin, Yutong and Cao, Yue and Hu, Han and Wei, Yixuan and Zhang, Zheng and Lin, Stephen and Guo, Baining},
  booktitle={Proceedings of the IEEE/CVF international conference on computer vision},
  pages={10012--10022},
  year={2021}
}

@article{ho2020denoising,
  title={Denoising diffusion probabilistic models},
  author={Ho, Jonathan and Jain, Ajay and Abbeel, Pieter},
  journal={Advances in neural information processing systems},
  volume={33},
  pages={6840--6851},
  year={2020}
}

@article{song2020denoising,
  title={Denoising diffusion implicit models},
  author={Song, Jiaming and Meng, Chenlin and Ermon, Stefano},
  journal={arXiv preprint arXiv:2010.02502},
  year={2020}
}

@inproceedings{rombach2022high,
  title={High-resolution image synthesis with latent diffusion models},
  author={Rombach, Robin and Blattmann, Andreas and Lorenz, Dominik and Esser, Patrick and Ommer, Bj{\"o}rn},
  booktitle={Proceedings of the IEEE/CVF conference on computer vision and pattern recognition},
  pages={10684--10695},
  year={2022}
}

@inproceedings{
lipman2023flow,
title={Flow Matching for Generative Modeling},
author={Yaron Lipman and Ricky T. Q. Chen and Heli Ben-Hamu and Maximilian Nickel and Matthew Le},
booktitle={The Eleventh International Conference on Learning Representations },
year={2023},
url={https://openreview.net/forum?id=PqvMRDCJT9t}
}

@inproceedings{ronneberger2015u,
  title={U-net: Convolutional networks for biomedical image segmentation},
  author={Ronneberger, Olaf and Fischer, Philipp and Brox, Thomas},
  booktitle={International Conference on Medical image computing and computer-assisted intervention},
  pages={234--241},
  year={2015},
  organization={Springer}
}

@article{karras2017progressive,
  title={Progressive growing of gans for improved quality, stability, and variation},
  author={Karras, Tero and Aila, Timo and Laine, Samuli and Lehtinen, Jaakko},
  journal={arXiv preprint arXiv:1710.10196},
  year={2017}
}

@article{morris2010control,
  title={Control of systems governed by partial differential equations},
  author={Morris, Kirsten},
  journal={The Control Systems Handbook},
  year={2018},
  publisher={CRC press}
}

@book{foias1996robust,
  title={Robust control of infinite dimensional systems: frequency domain methods},
  author={Foias, Ciprian and {\"O}zbay, Hitay and Tannenbaum, Allen},
  year={1996},
  publisher={Springer}
}

@misc{tay2022efficienttransformerssurvey,
      title={Efficient Transformers: A Survey}, 
      author={Yi Tay and Mostafa Dehghani and Dara Bahri and Donald Metzler},
      year={2022},
      eprint={2009.06732},
      archivePrefix={arXiv},
      primaryClass={cs.LG},
      url={https://arxiv.org/abs/2009.06732}, 
}

@inproceedings{
  dosovitskiy2020image,
  title={An Image is Worth 16x16 Words: Transformers for Image Recognition at Scale},
  author={Alexey Dosovitskiy and Lucas Beyer and Alexander Kolesnikov and Dirk Weissenborn and Xiaohua Zhai and Thomas Unterthiner and Mostafa Dehghani and Matthias Minderer and Georg Heigold and Sylvain Gelly and Jakob Uszkoreit and Neil Houlsby},
  booktitle={International Conference on Learning Representations (ICLR)},
  year={2021},
  url={https://openreview.net/forum?id=YicbFdNTTy}
}

@inproceedings{gu2021efficiently,
  title={Efficiently Modeling Long Sequences with Structured State Spaces},
  author={Albert Gu and Karan Goel and Christopher R{\'e}},
  booktitle={International Conference on Learning Representations (ICLR)},
  year={2022},
}

@misc{gu2023mamba,
  author       = {Gu, Albert and Dao, Tri},
  title        = {Mamba: Linear-Time Sequence Modeling with Selective State Spaces},
  year         = {2023},
  howpublished = {arXiv preprint arXiv:2312.00752}
}

@misc{zhu2024vision,
  author       = {Zhu, Lianghui and Liao, Bencheng and Zhang, Qian and Wang, Xinlong and Liu, Wenyu and Wang, Xinggang},

  title        = {Vision Mamba: Efficient Visual Representation Learning with Bidirectional State Space Model},
  year         = {2024},
  howpublished = {International Conference on Machine Learning}
}

@misc{krizhevsky2009learning,
  title={Learning multiple layers of features from tiny images.},
  author={Krizhevsky, Alex},
  year={2009}
}

@inproceedings{deng2009imagenet,
  title={ImageNet: A large-scale hierarchical image database},
  author={Deng, Jia and Dong, Wei and Socher, Richard and Li, Li-Jia and Li, Kai and Fei-Fei, Li},
  booktitle={2009 IEEE conference on computer vision and pattern recognition},
  pages={248--255},
  year={2009},
  organization={IEEE}
}

@inproceedings{Krizhevsky2012Imagenet,
  title     = {ImageNet Classification with Deep Convolutional Neural Networks},
  author    = {Krizhevsky, Alex and Sutskever, Ilya and Hinton, Geoffrey E.},
  booktitle = {Advances in Neural Information Processing Systems},
  year      = {2012},
}
\bibliographystyle{plainnat}



\appendix

\section{Additional Related Work}
\label{sec:additional_related_work}
We discuss additional related work that is relevant to our paper.

\citep{agarwal2023spectralsm} introduced spectral state space models that employ spectral filtering for long-range sequence dependencies, while \citep{fu2023simpleconv} proposed hardware-efficient long convolutions for sequence modeling. Both of these works remain in the sequence modeling setting, whereas our approach generalizes state space methods to multi-dimensional spatial domains via PDE operators.

In the generative modeling domain, \citep{goel2022rawaudio} explored the use of state-space models for audio generation, and \citep{shi2023multiresmem} developed a multiresolution convolutional memory approach for capturing long-range dependencies. These models highlight the applicability of SSM-like architectures to generative tasks, but they are constrained to one-dimensional or heuristic extensions of sequential structures. By contrast, PDE-SSM provides a principled extension to higher-dimensional spatial data and integrates directly into diffusion transformers.

\citep{smith2023simplessm} introduced simplified state space layers that offer efficient implementations of SSMs and serve as alternative baselines, and \citep{zhang2023discretessm} proposed discrete state space formulations for time series. Both approaches focus on efficiency and simplicity in the temporal domain, whereas our work emphasizes spatial generalization and non-locality through PDE formulations. Finally, \citep{tay2021lra} proposed the Long Range Arena benchmark for evaluating the scalability of models on long-range dependencies. While prior work is primarily evaluated on sequence tasks, PDE-SSM demonstrates scalability on image generation benchmarks, showcasing the benefits of our PDE-based perspective for vision applications.

Recent works investigate alternatives to attention within diffusion models. \citep{Teng2024DiM} replace attention with Mamba-based selective state space layers to accelerate high-resolution diffusion, but their formulation operates on flattened 1D token sequences. \citep{Qin2022HighlyAccurateDichotomousImageSegmentation} introduce a dichotomous refinement mechanism for segmentation, emphasizing the importance of strong spatial priors, an idea aligned with our PDE-based inductive bias. Finally, \citep{yan2023diffusion} propose DiffuSSM, which replaces attention with state-space layers inside diffusion models; however, DiffuSSM adapts 1D SSMs to images through sequential scanning rather than modeling true multidimensional structure. In contrast, PDE-SSM provides a learnable, physically grounded PDE operator that enables global non-local mixing with an efficient closed-form spectral implementation.
Fourier Neural Operators (FNOs) and spectral convolutions also operate in the frequency domain, but typically learn unconstrained complex-valued Fourier weights, often on a truncated set of low-frequency modes. In contrast, PDE-SSM uses a spectral symbol derived from the closed-form Green's function of a learnable convection--diffusion--reaction PDE. This parameterization naturally captures directed transport through convection and global amplification or suppression through reaction.

\section{Datasets and Experimental Settings}
\label{sec:datasets_experimental_settings}
We elaborate on our datasets and experimental settings.

\paragraph{CIFAR-10}
CIFAR-10 \citep{krizhevsky2009learning} is a canonical benchmark in computer vision, composed of 60,000 color images at a resolution of $32 \times 32$ pixels. The dataset is evenly divided into 10 object categories, including animals (e.g., birds, cats, dogs) and vehicles (e.g., airplanes, automobiles, ships). With 50,000 training images and 10,000 test images.

\paragraph{ImageNet$64$}
ImageNet64 \citep{deng2009imagenet} is a downsampled variant of the large-scale ImageNet dataset, where each image is resized to $64 \times 64$ pixels. It retains the rich diversity of the original ImageNet with over 1,000 categories spanning natural objects, animals, and everyday items. Despite the reduced resolution.

\paragraph{ImageNet$256$}
ImageNet \citep{Krizhevsky2012Imagenet} is a large-scale visual recognition benchmark containing over 1.2 million training images across 1,000 object categories. The dataset spans a wide range of natural objects, animals, and everyday scenes.

\paragraph{CelebA-HQ}
CelebA-HQ \citep{karras2017progressive} is a high-quality refinement of the CelebA dataset. The CelebA-HQ64 subset contains 30,000 face images at $64 \times 64$ resolution, emphasizing attributes such as hair style, facial expression, and background context.

\paragraph{LSUN-Churches}
The LSUN dataset \citep{yu2015lsun} contains millions of large-scale images categorized into different scene types. We focus on the LSUN-Churches subset, which consists of over 125,000 images of church images captured under diverse lighting conditions, architectural styles, and viewpoints. For our experiments, we use downsampled $64 \times 64$ versions, making the dataset suitable for structure-centric generative tasks that demand global spatial coherence.

\paragraph{Oxford-Flowers}
The Oxford 102 Flowers dataset \citep{nilsback2008automated} consists of 8,189 images across 102 flower categories, with substantial intra-class variation in scale, pose, and background. For this dataset, we report the FID score on the same number of images as the dataset size.

Our experimental setting and details are summarized in \Cref{tab:experimental_settings}.


\begin{table}[h]
\centering
\caption{Experimental settings and hyperparameters for different experiments.}
\resizebox{\textwidth}{!}{%
\begin{tabular}{lccccc}
\toprule
 & CIFAR-10 & ImageNet & CelebA-HQ & LSUN-Churches & Oxford-Flowers \\ \midrule
Max iter & 400,000 & 400,000 & 400,000 & 400,000 & 400,000 \\
Learning rate & $5 \times 10^{-4}$ & $5 \times 10^{-4}$ & $5 \times 10^{-4}$ & $5 \times 10^{-4}$ &  $5 \times 10^{-4}$ \\
Hidden size & 384 & 384 & 384 & 512 & 512 \\
Batch size & 64 & 64 & 64 & 32 & 32 \\
Image size & 32 & 64 & 64 & 64 & 64 \\
Layers & 12 & 12 & 12 & 12 & 12 \\
DiT Attention heads & 6 & 6 & 6 & 6 & 6 \\
\bottomrule
\end{tabular}
}

\label{tab:experimental_settings}
\end{table}

\FloatBarrier

\section{SSM Maps}
\label{sec:ssm_maps}

\begin{figure}[h]
    \centering
    \begin{subfigure}[t]{0.24\linewidth}
        \includegraphics[width=\linewidth]{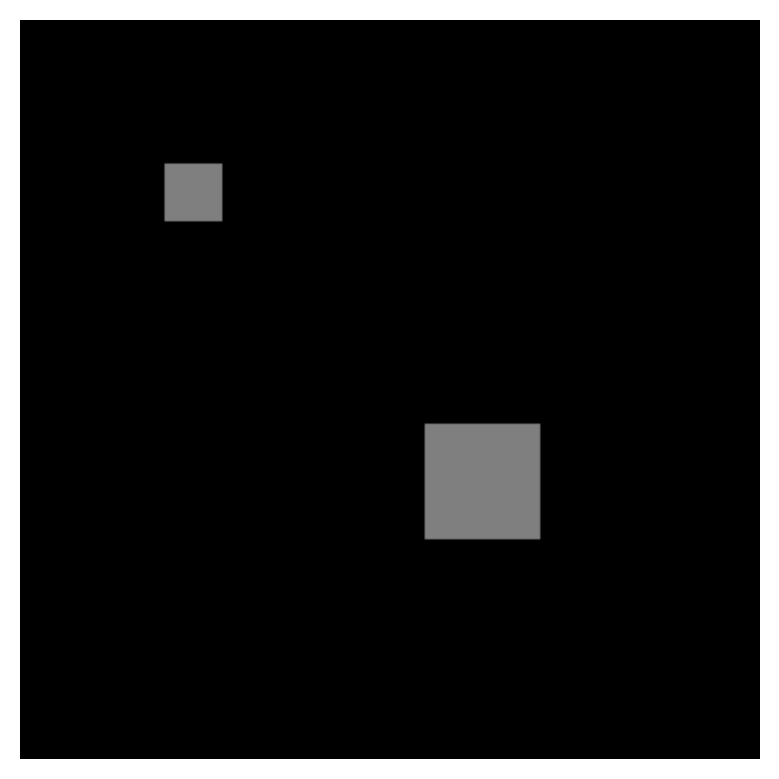}
        \caption{Original Simple}
        \label{fig:sub1}
    \end{subfigure}\hfill
    \begin{subfigure}[t]{0.24\linewidth}
        \includegraphics[width=\linewidth]{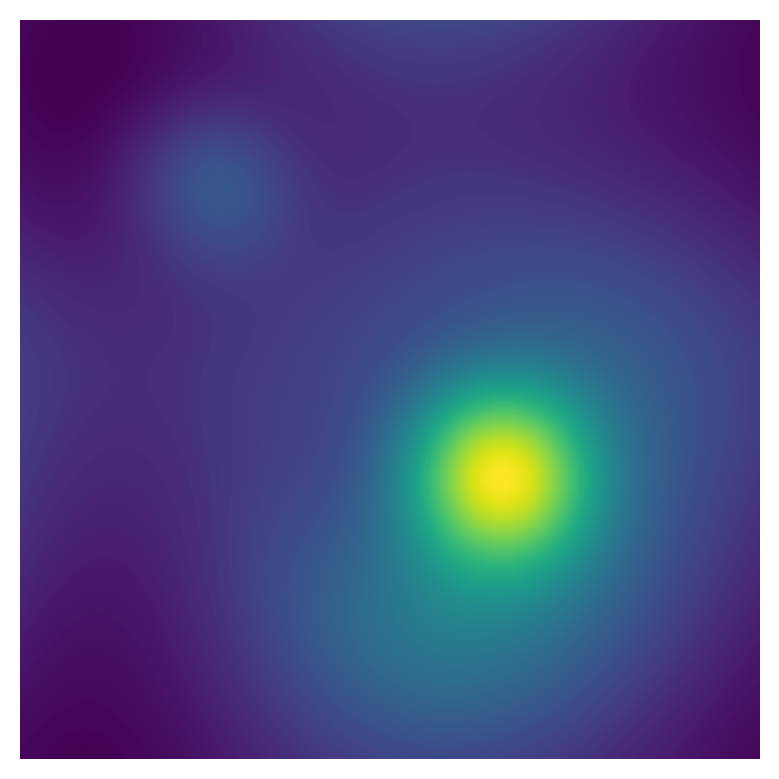}
        \caption{SSM Map - Simple}
        \label{fig:sub2}
    \end{subfigure}\hfill
    \begin{subfigure}[t]{0.24\linewidth}
        \includegraphics[width=\linewidth]{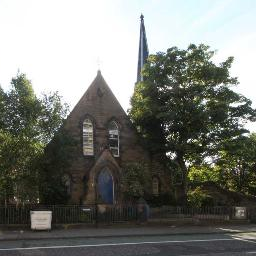}
        \caption{Original Church}
        \label{fig:sub3}
    \end{subfigure}\hfill
    \begin{subfigure}[t]{0.24\linewidth}
        \includegraphics[width=\linewidth]{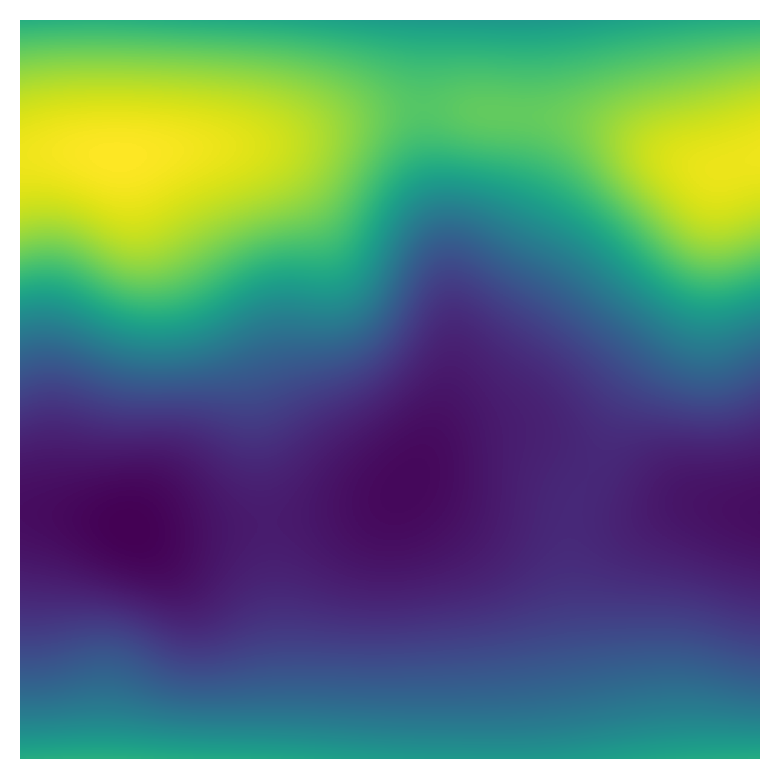}
        \caption{SSM Map - Church}
        \label{fig:sub4}
    \end{subfigure}
    \caption{SSM maps visualization: \textbf{(a)} real simple image; \textbf{(b)} simple image SSM map; \textbf{(c)} real church image; \textbf{(d)} church SSM map.}
    \label{fig:ssm_map}
\end{figure}

To better understand the behavior of our state-space layers, we visualize the SSM maps produced by our model. These maps highlight the global structures that the SSM captures across different images, demonstrating how information propagates over large spatial extents.
\Cref{fig:ssm_map} shows the SSM maps alongside their corresponding input images. In both the simple synthetic example and the real LSUN-Church \citep{yu2015lsun} image, the SSM map reflects the underlying object. This illustrates the strong global bias of the SSM mechanism, which aggregates information beyond local neighborhoods and responds to the broader shape and composition of the input.

\FloatBarrier

\section{Additional Timing Results}
\label{sec:timig}

To further demonstrate the efficiency of our SSM approach, we provide timing results for different model sizes and patch resolutions. In \Cref{tab:patch_sizes_compare}, we report the forward-pass time for the XL models across patch sizes, for an input size of 256 and batch size 4. As the patch size decreases, our method shows increasingly significant speedups, with especially large gains at the smallest patch resolution. \\
In \Cref{tab:model_size_compare}, we show that even at a fixed small patch size of 2, the SSM-based variants consistently outperform their DiT counterparts across all model scales.

We additionally report end-to-end inference and training performance across multiple resolutions and channel widths in \Cref{tab:inference_profile,tab:training_profile}, including peak memory and throughput. We also isolate the cost of the spatial mixing operators in \Cref{tab:operator_profile}. As the resolution increases, PDE-SSM-DiT exhibits increasingly favorable throughput relative to DiT.

\begin{table}[h]
\centering
\caption{Iteration time vs.\ patch size for DiT-XL and PDE-SSM-DiT-XL models.}
\label{tab:patch_sizes_compare}
\begin{tabular}{lcc}
\toprule
\textbf{Model} & \textbf{Patch Size} & \textbf{Time (s)} \\
\midrule
DiT-XL   & 8 & 0.1288 \\
DiT-XL   & 4 & 0.9633 \\
DiT-XL   & 2 & 10.4166 \\
\midrule
PDE-SSM-DiT-XL(Ours) & 8 & 0.1509 \\
PDE-SSM-DiT-XL(Ours) & 4 & 0.6315 \\
PDE-SSM-DiT-XL(Ours) & 2 & 2.3376 \\
\bottomrule
\end{tabular}
\end{table}

\begin{table}[h]
\centering
\caption{Iteration time for DiT and PDE-SSM-DiT models.}
\label{tab:model_size_compare}
\begin{tabular}{lcc}
\toprule
\textbf{Model} & \textbf{Variant} & \textbf{Time (s)} \\
\midrule
DiT & XL & 10.4162 \\
DiT & L  & 5.9323 \\
DiT & B  & 2.1223 \\
DiT & S  & 0.9863 \\
\midrule
PDE-SSM-DiT(Ours) & XL & 2.3435 \\
PDE-SSM-DiT(Ours) & L  & 1.6020 \\
PDE-SSM-DiT(Ours) & B  & 0.4773 \\
PDE-SSM-DiT(Ours) & S  & 0.1519 \\
\bottomrule
\end{tabular}
\end{table}

\begin{table*}[t]
\centering
\caption{Inference profiling across resolutions and channel widths.}
\label{tab:inference_profile}
{
\small
\begin{tabular}{cccccc}
\toprule
Resolution & Width & Architecture & Time (s) & Peak Mem. (GB) & Throughput (img/s) \\
\midrule
64  & 384 & DiT         & $1.179\times10^{-2}$ & 0.22 & 339.4 \\
64  & 384 & PDE-SSM-DiT & $1.112\times10^{-2}$ & 0.21 & 359.6 \\
64  & 768 & DiT         & $2.774\times10^{-2}$ & 0.69 & 144.2 \\
64  & 768 & PDE-SSM-DiT & $2.745\times10^{-2}$ & 0.68 & 145.7 \\
128 & 384 & DiT         & $8.228\times10^{-2}$ & 0.44 & 48.6 \\
128 & 384 & PDE-SSM-DiT & $3.186\times10^{-2}$ & 0.46 & 125.6 \\
128 & 768 & DiT         & $1.939\times10^{-1}$ & 1.11 & 20.6 \\
128 & 768 & PDE-SSM-DiT & $1.197\times10^{-1}$ & 1.36 & 33.4 \\
256 & 384 & DiT         & $9.907\times10^{-1}$ & 1.29 & 4.0 \\
256 & 384 & PDE-SSM-DiT & $1.517\times10^{-1}$ & 1.46 & 26.4 \\
256 & 768 & DiT         & $2.129\times10^{0}$               & 2.81 & 1.9 \\
256 & 768 & PDE-SSM-DiT & $4.790\times10^{-1}$ & 4.07 & 8.3 \\
\bottomrule
\end{tabular}
}
\end{table*}

\begin{table*}[t]
\centering
\caption{Training profiling, including the forward pass, backward pass, and optimizer step.}
\label{tab:training_profile}
{
\small
\begin{tabular}{cccccc}
\toprule
Resolution & Width & Architecture & Time (s) & Peak Mem. (GB) & Throughput (img/s) \\
\midrule
64  & 384 & DiT         & $3.905\times10^{-2}$ & 1.98  & 102.5 \\
64  & 384 & PDE-SSM-DiT & $3.206\times10^{-2}$ & 2.33  & 124.8 \\
64  & 768 & DiT         & $9.794\times10^{-2}$ & 4.74  & 40.8 \\
64  & 768 & PDE-SSM-DiT & $9.315\times10^{-2}$ & 6.54  & 42.9 \\
128 & 384 & DiT         & $3.143\times10^{-1}$ & 6.66  & 12.7 \\
128 & 384 & PDE-SSM-DiT & $9.647\times10^{-2}$ & 8.10  & 41.5 \\
128 & 768 & DiT         & $7.032\times10^{-1}$ & 14.08 & 5.7 \\
128 & 768 & PDE-SSM-DiT & $3.797\times10^{-1}$ & 21.84 & 10.5 \\
\bottomrule
\end{tabular}
}
\end{table*}

\begin{table*}[t]
\centering
\caption{Profiling of the isolated spatial mixing operator.}
\label{tab:operator_profile}
{
\small
\begin{tabular}{cccccc}
\toprule
Resolution & Width & Operator & Time (s) & Peak Mem. (GB) & Throughput (img/s) \\
\midrule
64  & 384 & Attention & $4.681\times10^{-4}$ & 0.06 & 8546.1 \\
64  & 384 & PDE-SSM   & $4.426\times10^{-4}$ & 0.08 & 9037.1 \\
64  & 768 & Attention & $1.063\times10^{-3}$ & 0.10 & 3761.9 \\
64  & 768 & PDE-SSM   & $8.945\times10^{-4}$ & 0.24 & 4471.9 \\
128 & 384 & Attention & $5.084\times10^{-3}$ & 0.17 & 786.8 \\
128 & 384 & PDE-SSM   & $9.018\times10^{-4}$ & 0.25 & 4435.5 \\
128 & 768 & Attention & $1.082\times10^{-2}$ & 0.33 & 369.6 \\
128 & 768 & PDE-SSM   & $4.482\times10^{-3}$ & 0.87 & 892.5 \\
256 & 384 & Attention & $7.423\times10^{-2}$ & 0.62 & 53.9 \\
256 & 384 & PDE-SSM   & $4.508\times10^{-3}$ & 0.94 & 887.4 \\
256 & 768 & Attention & $1.531\times10^{-1}$ & 1.24 & 26.1 \\
256 & 768 & PDE-SSM   & $1.752\times10^{-2}$ & 3.39 & 228.3 \\
\bottomrule
\end{tabular}
}
\end{table*}

\section{Additional ImageNet Results}
\label{app:imagenet}

For ImageNet64, we additionally report the individual results across three random seeds in Table~\ref{tab:imagenet64_seeds}. PDE-SSM-DiT achieves the lowest mean FID and the smallest standard deviation, indicating that the improvement over the DiT and U-Net baselines is consistent across runs.

\begin{table}[t]
\centering
\caption{ImageNet64 FID across three random seeds.}
\label{tab:imagenet64_seeds}

\begin{tabular}{lcccc}
\toprule
Method & Run 1 & Run 2 & Run 3 & Mean $\pm$ Std \\
\midrule
DiT & 22.6 & 23.2 & 22.9 & $22.9 \pm 0.30$ \\
U-Net & 21.7 & 21.8 & 23.1 & $22.2 \pm 0.78$ \\
PDE-SSM-DiT & 22.4 & 22.0 & 21.9 & $\mathbf{22.1 \pm 0.26}$ \\
\bottomrule
\end{tabular}

\end{table}

We report results for our PDE-SSM method on ImageNet$256$, summarized in \Cref{tab:imagenet}. 
We use the DiT-S/2 architecture and its equivalent PDE-SSM counterpart, trained for 300k iterations, and evaluate all models using classifier-free guidance.
As shown in the table, our method achieves performance comparable to the DiT baseline on the challenging ImageNet dataset, while offering improved modeling efficiency through structured state-space parameterization.

\begin{table*}[h]
\centering
\caption{Imagenet256 results.}
\begin{tabular}{lcccccc}
\hline
Model &  FID$\downarrow$ & sFID$\downarrow$ & IS$\uparrow$ & Precision $\uparrow$ & Recall $\uparrow$ \\
\hline
DiT & 48.44 & 9.04 & 30.78 & 0.45 & 0.53 \\
PDE-SSM-DiT-S-2 (Ours) & 49.37 & 11.21 & 34.09 & 0.47 &  0.48 \\
\hline
\end{tabular}
\label{tab:imagenet}
\end{table*}

\begin{figure*}[h]
    \centering
    \begin{subfigure}[t]{0.49\linewidth}        \includegraphics[width=\linewidth]{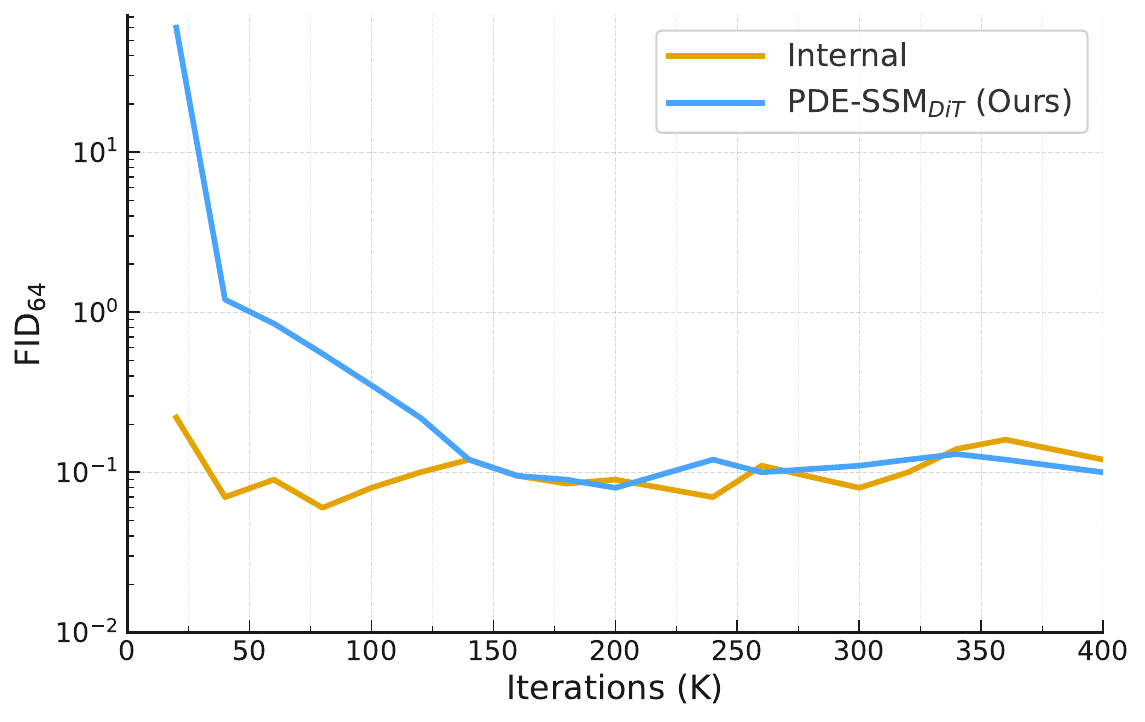}
        \caption{PDE-SSM-DiT \texttt{FID64} vs.\ internal baseline.}
        \label{fig:imagenet64_fid64-a}
    \end{subfigure}\hfill
    \begin{subfigure}[t]{0.49\linewidth}
        \includegraphics[width=\linewidth]{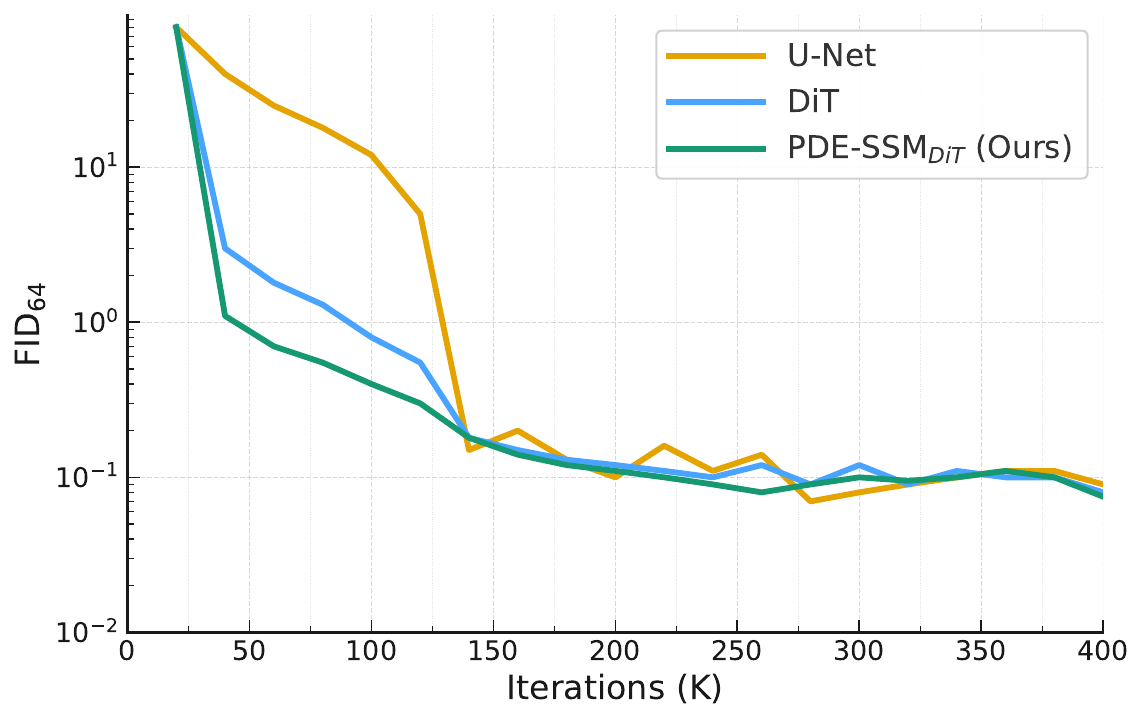}

                \caption{\texttt{FID64} across DiT, PDE-SSM-DiT, and U-Net.}
        \label{fig:imagenet64_fid64-b}
    \end{subfigure}
    \caption{ImageNet$64$ training. (a) The achieved FID score is consistent with the internal FID score. (b) All methods converge at a similar rate and to an FID score that is similar. }
    \label{fig:imagenet64_fid64}
\end{figure*}

Figure~\ref{fig:imagenet64_fid64} illustrates that DiT and PDE-SSM-DiT exhibit similar convergence behavior (panel b) and reach FID scores comparable to the dataset’s own internal FID (panel a). Here, internal FID refers to the baseline obtained by comparing the dataset against itself, serving as a reference for the best achievable score.

\section{Additional Samples}
\label{app:celcahq_samples}
\begin{figure*}[t]
    \centering
    \includegraphics[width=\textwidth]{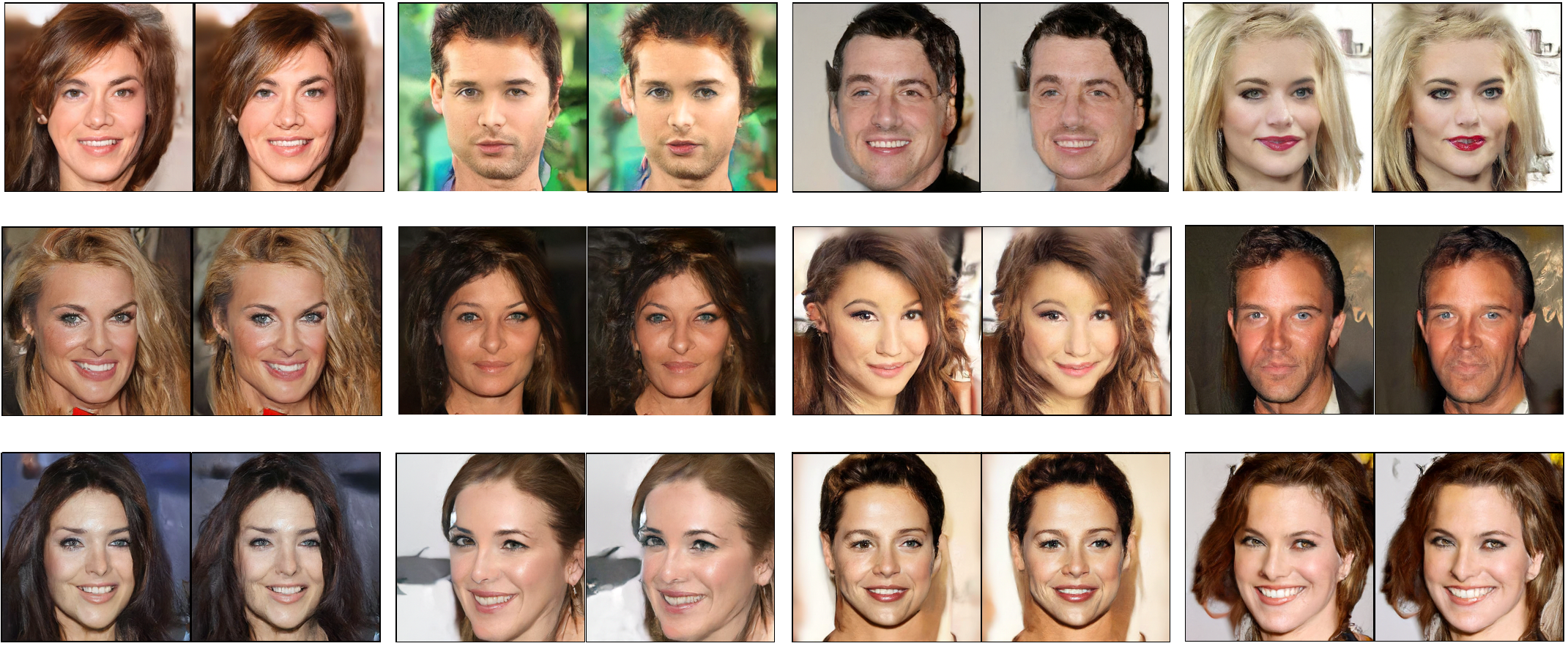}
    \caption{Sample celebrity faces generated by PDE-SSM-DiT (left) and DiT (right), trained on CelebA-HQ 256.}
    \label{fig:celebahq256}
\end{figure*}

We present qualitative results for PDE-SSM-DiT and DiT on CelebA-HQ 256 in \Cref{fig:celebahq256}, where the two models exhibit similar visual quality.

\section{Broader Impact}
\label{app:broader_impact}
While our work on PDE-SSM primarily presents a foundational architectural contribution that improves the efficiency and scaling of vision models, we recognize its broader societal implications. By reducing the computational bottleneck of spatial mixing from $\mathcal{O}(N^2)$ to $\mathcal{O}(N \log N)$ via spectral methods, our approach lowers the barrier to entry for training high-resolution generative models and helps mitigate the carbon footprint associated with large-scale AI research. Conversely, increasing the accessibility and efficiency of powerful generative architectures like Diffusion Transformers can inadvertently facilitate the scalable creation of malicious synthetic media, such as deepfakes used for disinformation or fraud. Mitigating these dual-use risks requires concerted downstream efforts, and we strongly encourage practitioners deploying PDE-SSM-based models to implement robust safeguards, including rigorous dataset curation, post-generation safety classifiers, and cryptographic watermarking to provenance AI-generated content.




\end{document}